\documentclass[11pt]{article}

\usepackage[
  shownumpages,          
  citingstyle=numbers,     
  bibliostyle=plainnat,    
  bibfile=references,      
]{brainlab}

\usepackage{shortcuts}

\usepackage{lipsum}

\setbrainmeta{
  title={Rethinking the Role of Tensor Decompositions \\in Post-Training LLM Compression},
  authors={Artur Zagitov\equalcontrib, Alexander Miasnikov\equalcontrib, Maxim Krutikov, Vladimir Aletov, Gleb Molodtsov, Nail Bashirov, Artem Tsedenov, Aleksandr Beznosikov
  },
  affiliations={
    BRAIn Lab, {\scriptsize{Moscow, Russia}} \\
  },
  abstract={
    Post-training compression is essential for deploying large language models (LLMs) under tight resource constraints. Tensor decompositions have emerged as a promising direction, offering compact parameterizations well suited to Transformer weight structures. However, existing studies evaluate these methods in narrow settings, leaving unclear whether tensorization is effective at large-scale deployment. We systematically evaluate tensor compression across dense and MoE architectures, establishing performance trade-offs grounded in both empirical analysis and theoretical analysis. We identify a fundamental mismatch between the shared subspaces assumed by tensor decompositions and the heterogeneous representations learned by modern LLMs, thereby delineating their practical limits and clarifying their viable role in large-scale deployment. The code is available at \url{https://github.com/brain-lab-research/TT-LLM}. 
  },
}

\begin{document}
\begin{mainpart}

\allowdisplaybreaks
\vspace{-8mm}
\section{Introduction}
\label{sec:intro}

In recent years, large language models (LLMs) have grown considerably in scale, 
increasing the storage and deployment costs and limiting their applicability 
in resource-constrained settings. Consequently, compression techniques are 
widely used to improve efficiency while preserving quality.

The primary goal of model compression is to reduce redundancy while preserving the model’s functional behavior. 
Standard approaches include pruning, which removes redundant components to decrease model size \citep{OBC_2022, LLM_pruner_2023, unreasonable_ineffectiveness_2024}; quantization, which represents weights and activations with lower-precision data types \citep{GPTQ_2022, QUIP_2023, AQLM_2024, AWQ_2024, harp}; and knowledge distillation (KD), which trains a compact student model to approximate the behavior of a larger teacher model \citep{HintonDistil_2015, GKD_2023}. 
However, achieving state-of-the-art performance with these techniques typically requires extensive fine-tuning or data-driven calibration.

A natural alternative is matrix and tensor decompositions, which are appealing due to their established theoretical background.
Methods in this category decompose a dense layer into a product of smaller factors, achieving parameter reduction while approximately maintaining the functionality of the original layer.
For a weight matrix $W \in \mathbb{R}^{m \times n}$, the SVD writes $W = U\Sigma V^\top$; truncating to the top $r$ singular values gives
\begin{equation}
\label{eq:svd}
\vspace{-1mm}
\widehat{W} = U_r\Sigma_r V_r^\top, 
\end{equation}
which by the Eckart--Young--Mirsky theorem~\citep{Eckart_Young_1936, Eckart_Young_Mirsky_1960} is globally optimal in any unitarily invariant norm.
Tensor decompositions extend this idea to multi-dimensional weight tensors in multi-head attention (MHA)~\citep{Attention_Is_All_You_Need_2017} and mixture of experts (MoE)~\citep{MoE_2022}: Tucker decomposition~\citep{tucker1966some, kolda2009tensor} approximates a
tensor $\mathcal{X} \in \mathbb{R}^{I_1 \times \cdots \times I_N}$ via a
compressed core tensor $\mathcal{G} \in \mathbb{R}^{R_1 \times \cdots \times R_N}$
and per-mode factor matrices $U^{(n)} \in \mathbb{R}^{I_n \times R_n}$:
\begin{equation}
    \label{eq:tucker}
    \vspace{-2mm}
    \mathcal{X} \approx \mathcal{G} \times_1 U^{(1)} \times_2 U^{(2)}
    \cdots \times_N U^{(N)},
\end{equation}
where $(R_1,\ldots,R_N)$ are the Tucker ranks.

Tensor Train (TT)~\citep{oseledets2011tensor} chains three-dimensional cores:
\begin{equation}
    \label{eq:tt}
    \vspace{-2mm}
    \mathcal{X}_{i_1,\ldots,i_N} \approx \!\!\sum\nolimits_{r_1,\ldots,r_{N-1}}\!\! \mathcal{G}^{(1)}_{1,i_1,r_1}\mathcal{G}^{(2)}_{r_1,i_2,r_2} \cdots \mathcal{G}^{(N)}_{r_{N-1},i_N,1},
\end{equation}
where $\mathcal{G}^{(n)} \in \mathbb{R}^{R_{n-1} \times I_n \times R_n}$, $R_0 = R_N = 1$, and $(R_1,\ldots,R_{N-1})$ are the TT ranks.


Despite their theoretical appeal, existing studies on tensor
decompositions~\citep{tensorllm2025,li2026lestd} report positive results under
evaluation protocols that do not reflect full-scale deployment constraints,
leaving the practical utility of these methods unclear in realistic settings. Therefore, we contribute the following:

$\bullet$ \textbf{Systematic empirical validation and deep diagnosis of matrix and tensor decompositions.} We conduct the first large-scale study of post-training tensor decompositions under realistic deployment conditions, covering dense and MoE architectures, evaluated on perplexity, downstream accuracy, and a lightweight LoRA repair baseline. Across all settings, tensor formats fail to outperform their matrix counterparts at matched compression ratios. We attribute this to a consistent failure mechanism, identified through perplexity, metrics and residual-stream geometry, and also providing an ablation of super-weight~\citep{superweight} restoration sensitivity to Frobenius-optimal truncation.

$\bullet$ \textbf{Partial theoretical explanation.} We formalize three chained obstructions that prevent any Frobenius-optimal tensor decomposition from serving as an effective compressor of Transformer weights without significant fine-tuning.

\section{Related Work}

\paragraph{Low-rank matrix compression.}
As mentioned in Section~\ref{sec:intro} truncated SVD (Eq.~\ref{eq:svd}) guarantees globally optimal low-rank approximation.


LASER~\citep{LASER_2024} applies truncated SVD directly to selected
weight matrices without retraining.
SliceGPT~\citep{slicegpt2024} exploits the rotational invariance of
RMSNorm-connected Transformers to absorb PCA bases into adjacent layers.
SVD-LLM~\citep{svdllm2025} improves rank selection via
truncation-aware data whitening and a layer-wise closed-form update.
Dobi-SVD~\citep{wang2025dobisvd} makes the truncation position
differentiable, optimising it on calibration data with respect to the
activation.
SoLA~\citep{huang2025sola} preserves a small set of high-activation-norm
FFN neurons in dense form and applies low-rank decomposition only to the
remainder.
FLAT-LLM~\citep{tian2025flatllm} applies head-wise PCA for attention
blocks and Nystr\"{o}m approximation for MLP blocks.
HASSLE-free~\citep{hasslefree2025} decomposes each weight into a sparse
plus low-rank component. COALA~\cite{parkina2026coala} addresses numerical instabilities inherent in context-aware low-rank compression by replacing explicit Gramm matrix formation and inversion with a stable QR-based projection, thereby avoiding the singularity issues that arise in prior methods such as SVD-LLM.

\paragraph{Tensor decomposition for LLMs.}
TensorLLM~\citep{tensorllm2025} and LeSTD~\citep{li2026lestd} apply Tucker
decomposition to multi-head attention projections, with LeSTD additionally
sparsifying the Tucker core for higher compression ratios.
TD-MoE~\citep{tdmoe2026} extends Tucker decomposition to sparse MoE models
by stacking expert weights into a three-dimensional tensor.
TRAWL~\citep{luo2024trawl} stacks related weight matrices across layers and
applies CP/Tucker-style decomposition as a training-free denoising
intervention.
A separate line of work uses TT for parameter-efficient fine-tuning
rather than post-training compression: LoRETTA~\citep{yang2024loretta}
introduces TT-based adapters, TT-LoRA~\citep{anjum2024ttlora} parameterizes
adaptation updates with TT cores, and AdaZeta~\citep{yang2024adazeta}
combines tensorized adapters with zeroth-order fine-tuning.

\begin{figure*}[!b] 
  \centering
  \vspace{-2mm}\includegraphics[width=0.9\textwidth]{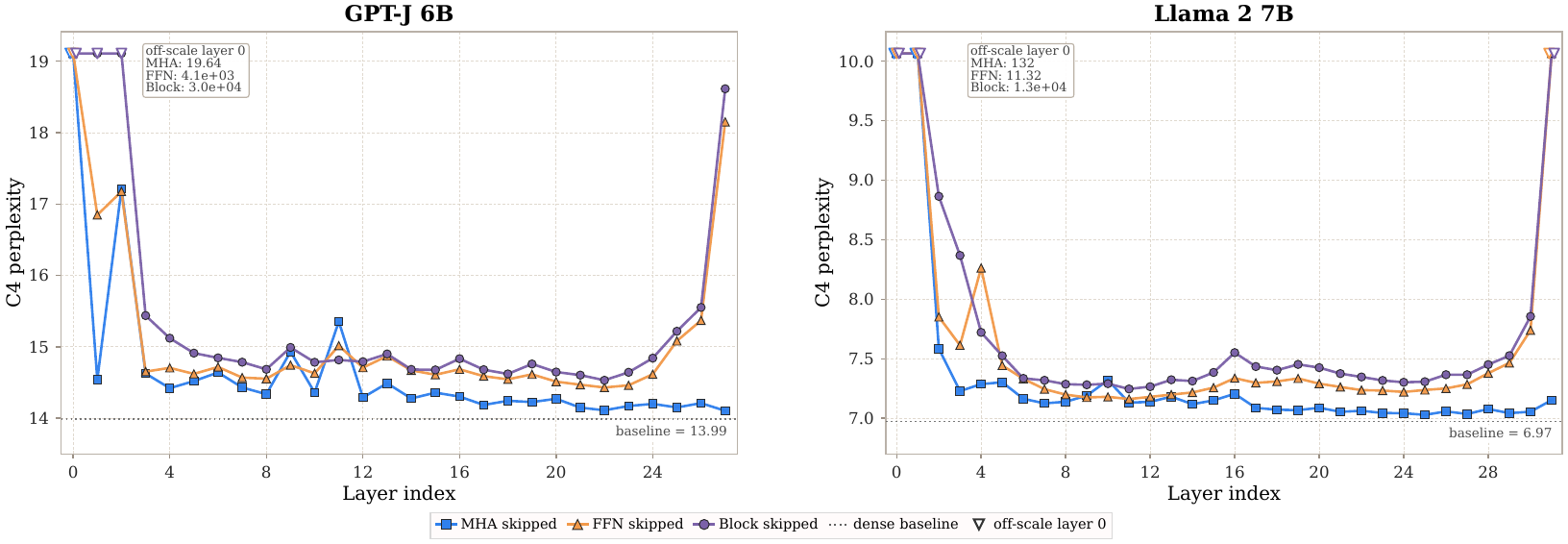}
  \caption{Perplexity (PPL) over pruning. Displays the final PPL when each layer is individually pruned.}
  \label{fig:full_width_graph}
  \vspace{-2mm}
\end{figure*}

\section{Compression Strategies}
\label{sec:compression}

\subsection{Pruning}

Pruning removes parameters or components that contribute least to the model’s behavior. It is done by exploiting the highly non-uniform parameter redundancy inherent in LLMs. Notably, both prior research \citep{unreasonable_ineffectiveness_2024} and our own experiments (Figure~\ref{fig:full_width_graph}) demonstrate that intermediate layers have relatively little impact on final quality and can tolerate aggressive pruning. This phenomenon arises because the earliest and latest layers disproportionately handle critical tasks, such as forming token representations and generating final predictions, while the intermediate layers play a comparatively smaller role.
However, while pruning effectively discards these middle-layer tails, it limits the maximum achievable compression ratio due to the dominant mass of the parameters intact. It does not compactly reparameterize the remaining structure, making it a na\"ive baseline for more structurally advanced decompositions.


\subsection{Matrix Decompositions}


A natural next step is matrix factorization, which compresses weights via low-rank reparameterization rather than deletion. For Feed-Forward Networks (FFNs), which contain the majority of model parameters \citep{FFNs_Are_Memory_2021}, we select and evaluate several matrix decompositions using the protocol described in Appendix~\ref{app:gptj_llama_tensor_protocol}.

\texttt{LASER}~\citep{LASER_2024} applies
Eq.~\ref{eq:svd} to individual FFN weight matrices without retraining or calibration data via selecting the layer, matrix type, and rank $r$ by grid search over downstream task accuracy. Rank reduction can improve performance on reasoning tasks, which is interpreted as a \emph{denoising} effect: low-singular-value directions are hypothesised to encode competing responses that suppress weakly learned facts. \texttt{LASER} applies no correction for truncation error, making it a natural training-free baseline.



\texttt{HASSLE-free}~\citep{hasslefree2025} method decomposes each weight matrix into a
sparse and a low-rank component, namely
\begin{equation}
    W \approx S + AB^{\top},
\end{equation}
minimizing the layer-wise activation reconstruction error $\|Wx - (S + AB^{\top})x\|$ via alternating minimization: $S$ is updated by thresholding the residual $W - AB^{\top}$, and $AB^{\top}$ by solving a least-squares problem on $W - S$. The sparse
component captures isolated high-magnitude entries that low-rank factors approximate poorly, while the low-rank component encodes the structured signal.

\begin{figure}[htbp]
  \centering
  \begin{minipage}[t]{0.48\textwidth}
  \texttt{SoLA}~\citep{huang2025sola} leverages deeper matrix structure partition with a dense subset of top-$k$ highest-activation-norm neurons, kept at full precision, and a compressed subset to which truncated SVD is applied. This avoids the disproportionately large errors that arise when high-norm neurons are low-rank approximated. Additionally, \texttt{SoLA} allocates ranks across layers via integer programming, minimizing total reconstruction error subject to a global parameter budget, concentrating capacity where truncation error is
most sensitive.

    \texttt{FLAT-LLM}~\citep{tian2025flatllm} applies head-wise PCA to attention blocks and Nystr\"{o}m approximation to MLP blocks, computing truncation bases from calibration activations. Ranks are distributed across decoder layers via a greedy sensitivity-based redistribution strategy, allocating the global rank budget to layers where truncation causes the largest reconstruction error.
  \end{minipage}%
  \hfill%
  \begin{minipage}[t]{0.48\textwidth}
    \vspace{0pt}
    \centering
    \includegraphics[width=\linewidth]{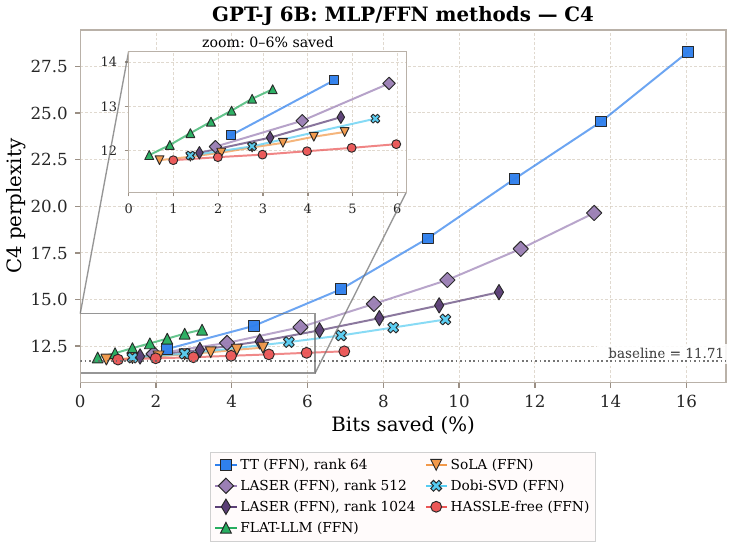}
    \captionof{figure}{FFN compression on GPT-J~6B. All methods compress the
    same middle-to-late block ranges.}
    \label{fig:matrix_decomps}
  \end{minipage}
  \vspace{-1em}
\end{figure}

Figure~\ref{fig:matrix_decomps} reveals a clear trend: methods that employ
some form of fine-tuning or leverage additional calibration data (such as
\texttt{Dobi-SVD}, \texttt{HASSLE-free}, and \texttt{SoLA}) exhibit
substantially smaller performance degradation compared to methods relying
solely on matrix decomposition, such as \texttt{LASER} and
\texttt{Flat-LLM}.

Nevertheless, the highest compression ratio is achieved by LASER ($\approx 14\%$), yet with $+8$ PPL increase. This raises the natural questions which we adress in the Section~\ref{sec:td-dense}:

\begin{tcolorbox}
    \textbf{Q1}: \textit{Can we achieve a higher compression ratio without sacrificing model quality?}
\end{tcolorbox}

\begin{figure*}[t!]
  \centering
  \includegraphics[width=0.85\textwidth]{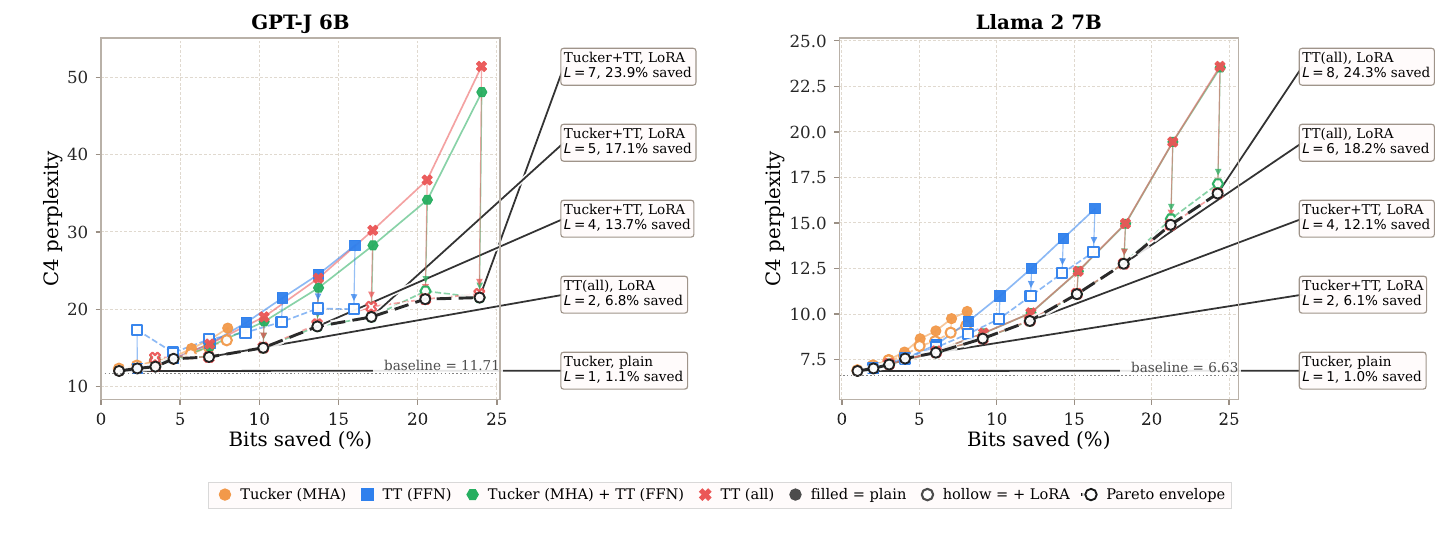}
  \vspace{-2mm}
  \caption{C4 perplexity versus bits saved, excluding embeddings, for GPT-J 6B and LLaMA 2 7B. Each point is one compression run, and $L$ denotes the number of consecutive compressed transformer blocks. Arrows connect each decomposition to its LoRA-repaired variant, and the Pareto frontier marks the best observed trade-offs.}
  \label{fig:c4_pareto}
  \vspace{-1em}
\end{figure*}

\subsection{Tensor decompositions for Dense Architectures}
\label{sec:td-dense}

It is well known that tensor decompositions allow exponential compression of a dense tensor with respect to its dimensionality. Methods such as TT-SVD~\citep{oseledets2011tensor} and HOSVD~\citep{hosvdopt} achieve quasi-optimal approximation error in the Frobenius norm, representing a tensor in TT and Tucker formats respectively. By employing these representations in operator format~\citep{tyrtyshnikov2003tensor, hackbusch2006low, oseledets2011tensor}, one can compress an FFN weight matrix by first
reshaping it into a tensor. Figure~\ref{fig:matrix_decomps} partially addresses \textbf{Q1} by comparing FFN-layer TT decomposition against matrix-based methods on GPT-J~6B. The results clearly show that TT decomposition performs worser than \texttt{LASER} in terms of PPL, yet achieving greater compression.

On the other hand, as demonstrated in \texttt{TensorLLM}~\citep{tensorllm2025}, the MHA layer admits a natural four-dimensional tensor representation that makes its low- dimensional structure explicit, motivating Tucker decomposition as the most structurally appropriate format in this setting. For FFN blocks, which account for the majority of model parameters, no such natural tensor structure exists; thus we are forced to apply TT decomposition to match the compression ratio of matrix methods and fully address the \textbf{Q1}. To maximize compression, in all experiments with TT decomposition we reshape the weight matrix into $12$ cores, since the smallest dimension in both models is $4096 = 2^{12}$.  The resulting decomposition, however, introduces non-negligible perplexity degradation; to partially recover quality, we apply a lightweight LoRA repair stage ($r=16$, trained on WikiText-2) as a parameter-efficient nanlogue for the calibration-aware weight decompositions used in matrix compression methods~\citep{wang2025dobisvd,svdllm2025}. We evaluate GPT-J~6B~\citep{gpt-j} and LLaMA~2~7B~\citep{touvron2023llama} under four compression schemes: Tucker on attention only (reproducing \texttt{TensorLLM}-style factorisation), TT on FFN only, Tucker+TT jointly, and TT on all projections, with a maximum Tucker rank of $64$, following the protocol detailed in Appendix~\ref{app:gptj_llama_tensor_protocol}. Additionally, we report further combinations of tensor decomposition methods in Appendix~\ref{app:detailed_gptj_llama_results}.



Figure~\ref{fig:c4_pareto} summarizes the results. Attention-only compression preserves perplexity but yields negligible size reduction; compressing FFN layers achieves higher compression at the cost of sharp quality degradation. LoRA repair partially recovers quality but does not resolve the fundamental trade-off. Despite structural alignment between tensor formats and LLM components, all methods exhibit poor compression-quality trade-offs at practical compression ratios, pointing to a deeper mismatch between standard tensor assumptions and learned LLM representations. We also provide trade-offs for WikiText-2 perplexity and macro LM-Eval accuracy drop in Appendix~\ref{app:quality_compression}.

\subsection{Tensor decompositions for Mixture-of-Experts}
\label{sec:td-moe}

\begin{wrapfigure}{r}{0.52\textwidth}
\vspace{-4mm}
\centering
\includegraphics[width=\linewidth]{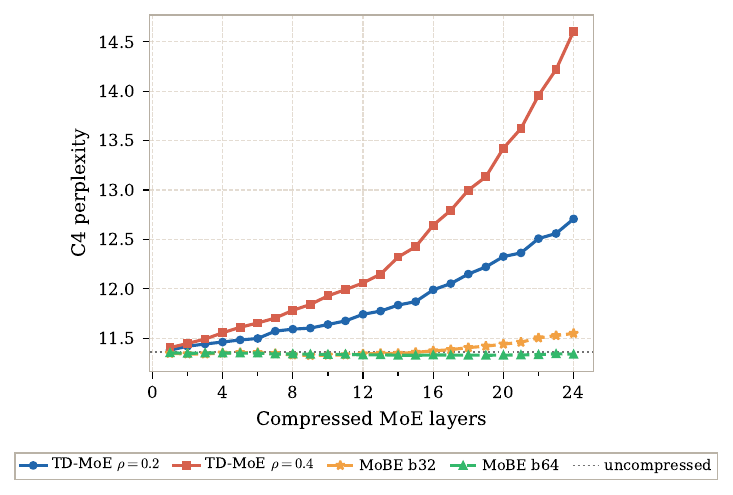}
\caption{C4 perplexity of \texttt{TD-MoE} and \texttt{MoBE} on Qwen3-30B-A3B model.}
\label{fig:moe_main}
\vspace{-6mm}
\end{wrapfigure}
MoE layers are an especially appealing target for tensorization. Since Switch Transformers \citep{switch}, MoE models have been trained under a trade-off between expert specialization and load balancing, often leading to overlapping representations across experts. This effect becomes even more pronounced in grouped designs \citep{moge,himoe}. Consequently, the expert dimension provides a natural tensor mode through which Tucker decomposition can exploit cross-expert redundancy, as proposed in \texttt{TD-MoE} \citep{tdmoe2026}. A complementary matrix-based approach is taken by \texttt{MoBE}~\citep{chen2025mobe}, which exploits cross-expert redundancy by factorizing each expert's up-/gate weight matrix as $W = AB$, where $A$ is expert-specific and $B$ is reparameterized as a linear combination of basis matrices $\{B_i\}$ shared across all experts within a layer, with the decomposition learned by minimizing the layer-wise reconstruction error. We test whether the structural alignment of tensor formats translates into better post-training compression on Qwen3-30B-A3B~\citep{qwen3technicalreport} (128 experts, 8 active) benchmarking \texttt{TD-MoE} against \texttt{MoBE} as a matrix-decomposition baseline. Additional experiments with GPT-OSS-20B~\citep{openai2025gptoss} can be find in Appendix~\ref{app:td_moe_gpt_oss}.
 
Following the protocol from Appendix~\ref{app:td_moe}, Figure~\ref{fig:moe_main} summarizes the comparison between \texttt{MoBE} and \texttt{TD-MoE}. Surprisingly, \texttt{TD-MoE} performs substantially worse than \texttt{MoBE}, despite both methods employing gradient-based optimization during compression. \texttt{MoBE} remains close to the original model across all tested compression ratios, whereas \texttt{TD-MoE} degrades significantly, suggesting that the structural alignment between the tensor format and the expert index does not by itself confer a compression advantage over matrix-based factorization with shared bases. Thus, Tucker decomposition for MoE yields substantially better results than for MHA, yet remains considerably weaker than the matrix-based method MoBE. This observation leads to the following question.

\begin{tcolorbox}
    \textbf{Q2}: \textit{What are the underlying reasons for such behaviour of tensor decompositions across the two considered architectures?}
\end{tcolorbox}

\paragraph{Expert mode is the binding constraint.}
A controlled ablation on GPT-OSS-20B at $\rho = 0.4$ compares two
rank allocations at equal storage:
\emph{preserve} fixes the expert-mode rank to $r_1 = K$ (all experts
remain distinct latent directions), while \emph{compress} reduces it
to $r_1 < K$ and re-spends the saved budget on the feature modes
(Appendix~\ref{app:td_moe_gpt_oss}, Fig.~\ref{fig:gpt_oss_preserve_vs_compress}).
\emph{compress} consistently underperforms \emph{preserve} across all
benchmarks, confirming that, at least for this model, expert directions are not interchangeable
and that the natural Tucker assumption ``experts live in a low-rank
subspace'' is the binding constraint.
 
 

\begin{figure*}[t]
  \centering
  \begin{subfigure}[t]{0.60\textwidth}
    \centering
    \includegraphics[width=\linewidth]{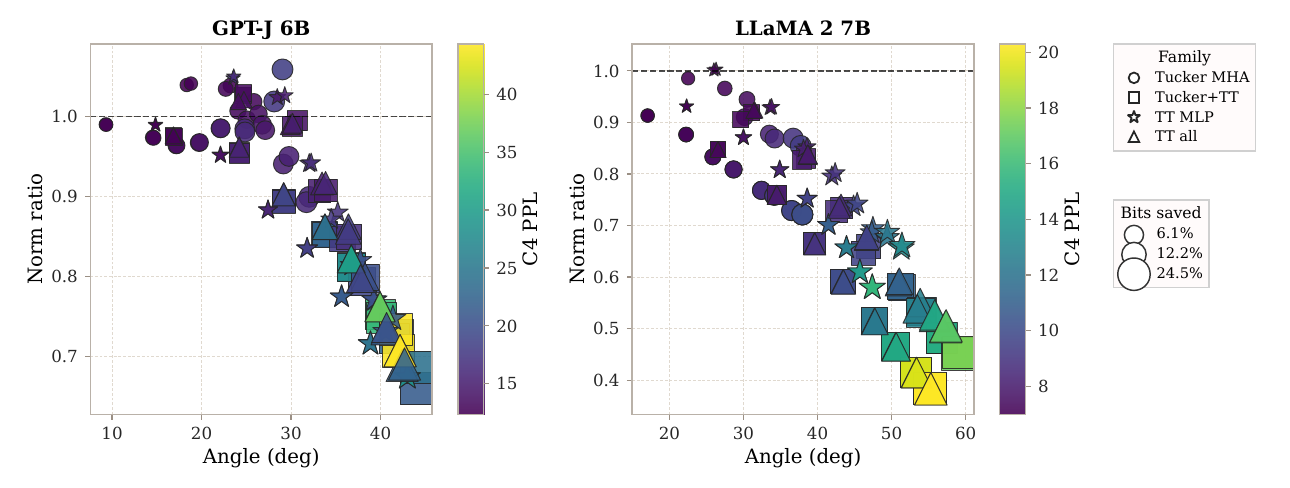}
    \caption{Activation geometry of dense-model versus compressed-model GPT-J~6B and LLaMA~2~7B.}
    \label{fig:activation_geometry_dense}
  \end{subfigure}
  \hfill
  \begin{subfigure}[t]{0.38\textwidth}
    \centering
    \includegraphics[width=\linewidth]{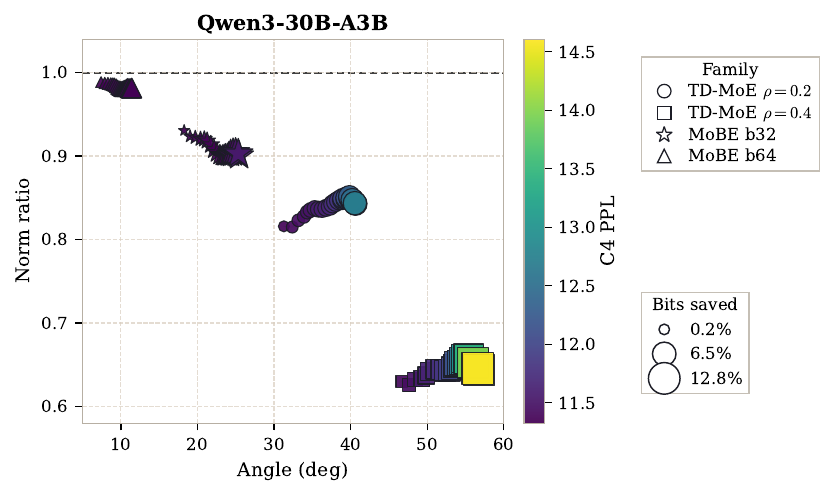}
    \caption{Activation geometry of \texttt{TD-MoE}
    for $\rho \in \{0.2, 0.4\}$ versus \texttt{MoBE} for Qwen3-30B-A3B.}
    \label{fig:activation_geometry_moe}
  \end{subfigure}
  \vspace{-0.5em}
  \caption{\textbf{Residual-stream activation geometry.}
  \textbf{(a)}~Under dense-model tensor compression.
  \textbf{(b)}~Under MoE tensor versus matrix compression.
  Each point is one compression run across the tested layer ranges.
  Color indicates C4 perplexity, marker shape indicates the decomposition family,
  and marker size indicates bits saved excluding embeddings.}
  \label{fig:activation_geometry}
\end{figure*}

\subsection{Structural Mismatch of Tensor Decompositions in LLMs}
\label{sec:mismatch}


To address \textbf{Q2}, we conduct an experiment, where for each compression run -- dense or MoE -- we compare residual-stream activations of the original model and of the compressed one after the last decomposed block. The diagnostic captures the accumulated effect of all decomposed layers up to the target compression ratio rather than the local error of a single layer. We quantify the deviation by the mean angle between dense and compressed activation vectors and by their norm ratio, averaged over evaluation tokens.

Figure~\ref{fig:activation_geometry} shows the same pattern for all models. Runs with low perplexity remain close to the dense-model trajectory in both direction and scale, whereas high-perplexity runs exhibit larger angular drift, norm shrinkage, or both. This suggests that tensorization degrades quality by progressively distorting the geometry of the residual stream, rather than merely increasing parameter reconstruction error. Thus, the main limitation is not the local expressivity of Tucker or TT decompositions, but their inability to preserve the heterogeneous representation geometry induced by LLMs without additional training. We explain this in the next section.

%





\subsection{Important Weights for Activation Restoration}
\label{sec:superweights}

\begin{figure*}[t]
  \centering
  \includegraphics[width=0.9\textwidth]{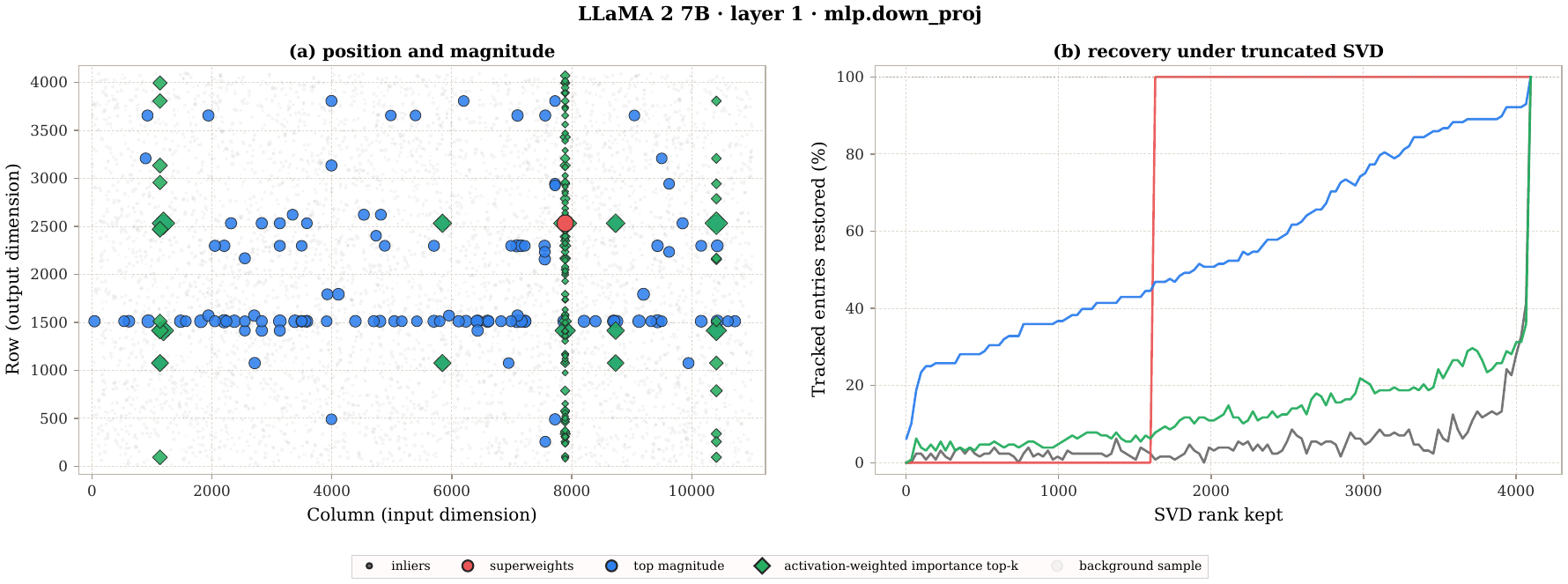}
  \vspace{-2mm}
  \caption{\textbf{(a)}~Positions of inliers, the known super-weight, top-magnitude entries,
    and activation-weighted top-$k$ coordinates in the weight matrix.
    \textbf{(b)}~Fraction of tracked entries recovered as a function of retained SVD rank.
    }
  \label{fig:superweigths}
  \vspace{-1em}
\end{figure*}

It is well-known that a small fraction of LLM parameters, termed super-weights and outliers~\citep{superweight}, are disproportionately critical to model quality: removing as few as a single scalar can raise perplexity by several orders. These parameters are not essentially the largest in magnitude, but occupy specific coordinates that dominate particular activation directions while carrying negligible Frobenius mass. Standard low-rank projections, which minimize a global Frobenius objective, therefore systematically discard them, producing severe spectral distortions at precisely the coordinates that matter most. Super weights further induce persistent large activation outliers across layers~\citep{sun2024massive,SinksAndSpikes_2026}, amplifying this effect under compression. 

To empirically validate this claim, we first observe that LoRA fine-tuning restores the original super-weight scalar with an absolute error of $10^{-4}$ on the LLaMA~2~7B model, whose super-weight resides at coordinates $(2533,\,7890)$ of the first MLP down-projection matrix~\citep{SinksAndSpikes_2026}. For a more systematic investigation, we apply \texttt{HASSLE-free} to this matrix: its sparse/low-rank decomposition combined with Hessian-based importance scoring recovers $95\%$ of the original uncompressed activation magnitude, providing a principled reference set of critical weight coordinates. We then apply truncated SVD at increasing ranks to the same matrix in order to determine the minimum rank required to individually restore the super-weight, the activation outliers, and the full set of \texttt{HASSLE-free}-selected coordinates. We also provide experiments demonstrating that pruninng of this layer collapses the model in Appendix~\ref{app:single_layer_decomposition}.

The results are presented in Figure~\ref{fig:superweigths}. As shown in the first panel, the column indices of the weights selected by \texttt{HASSLE-free} closely coincide with those of the identified outliers and super-weight coordinates. Furthermore, the second panel of Figure~\ref{fig:superweigths} reveals that restoring the super-weight alone requires a rank of at least $1500$, while recovering the full set of critical weights identified by \texttt{HASSLE-free} demands a rank approaching that of the original matrix. This result has a direct implication: decomposition-based methods that retain only the leading singular components systematically fail to capture the fine-grained structure of LLM weights, since faithful reconstruction requires $r \gg \nicefrac{(\min \{m, n\})}{2}$, far beyond the ranks used in practice. To isolate the role of super-weights, we repeat the same experiment on a layer that contains no super-weight and obtain qualitatively identical restoration results (Appendix~\ref{app:no-superweights}). This control experiment is particularly revealing: it demonstrates that super-weight restoration, while necessary, is not sufficient for activation recovery, and that the pathological sensitivity to rank truncation reflects a deeper structural property of LLM weight matrices rather than a localised outlier phenomenon. We therefore conclude that na\"{i}ve rank truncation is an inherently inadequate strategy for compressing LLM weights.

\section{Why Tensorization Fails: An Operator Analysis Perspective}
\label{sec:theory}

The empirical patterns such as sharp degradation under FFN tensorization in Fig.~\ref{fig:matrix_decomps}), residual-stream geometric drift (Fig.~\ref{fig:activation_geometry}), pathological sensitivity to rank truncation (Fig.~\ref{fig:superweigths}), and the consistent gap between \textsc{TD-MoE} and the matrix-based \textsc{MoBE} (Fig.~\ref{fig:moe_main}) point to a \textit{structural} rather than a methodological cause. We formalize this through two chained obstructions: Frobenius-optimal tensor decompositions (HOSVD, TT-SVD, Tucker) (i)~optimize a norm misaligned with operator-norm preservation, and (ii)~on the heavy-tailed spectra that Transformer weights exhibit, incur a spectral error exceeding the matrix-optimal error by a factor growing polynomially with layer width.

\paragraph{Setup.}
Let $W \in \mathbb{R}^{m \times n}$ be a layer weight and $\varphi : \mathbb{R}^{m \times n} \to \mathbb{R}^{I_1 \times \cdots \times I_d}$ a reshape with $\prod \limits_{\ell = 1}^d I_\ell = mn$ and $d \geq 2$. Set $\mathcal{T} = \varphi(W)$, let $\widehat{\mathcal{T}} \in \mathcal{M}_{\mathbf{r}}$, where $\mathcal{M}_{\mathbf{r}}$ is a fixed rank manifold, be a Frobenius-optimal low-rank approximation (Tucker rank $\mathbf{r}=(r_1,\dots,r_d)$ via HOSVD or TT rank $(r_1,\dots,r_{d-1})$ via TT-SVD), and $\widehat{W} = \varphi^{-1}(\widehat{\mathcal{T}})$. Singular values of the mode-$\ell$ unfolding $\mathcal{T}_{(\ell)}$ are $\sigma_{\ell,1} \geq \sigma_{\ell,2} \geq \dots$; we drop the mode subscript when clear.

\paragraph{Tensorization preserves Frobenius but not operator geometry.} Since $\varphi$ permutes entries, $\|\varphi(A)\|_F = \|A\|_F$ for all $A$. The tensor spectral (injective) norm,
\begin{equation}
\|\mathcal{T}\|_\sigma \;:=\; \sup_{\|u^{(\ell)}\|_2 = 1} \bigl\langle \mathcal{T},\, u^{(1)} \otimes \cdots \otimes u^{(d)} \bigr\rangle,
\label{eq:tensor-spectral}
\end{equation}
takes its supremum over rank-one unit tensors: a strict subset of the unit operator-norm ball when $d \geq 3$. Hence $\|\varphi(W)\|_\sigma \leq \|W\|_2$, with equality only when $\varphi(W)$ is rank-one. For $d \geq 3$ computing $\|\mathcal{T}\|_\sigma$ is computationally hard, and the two norms can be polynomially separated.

\begin{proposition} [\citealp{friedland2018nuclear}]
\label{prop:gap}
For $W \in \mathbb{R}^{m \times n}$ reshaped to order $d \geq 3$,
\begin{equation}
\frac{\|W\|_2}{\|\varphi(W)\|_\sigma} \;\in\; \Bigl[\,1,\; \sqrt{\min(m,n)}\,\Bigr],
\label{eq:gap-bound}
\end{equation}
and the upper bound is generically attained on matrices with delocalised singular vectors. 
\end{proposition}

Consequently, $\varphi$ is a Frobenius isometry but not an isometry between $(\mathbb{R}^{m \times n}, \|\cdot\|_2)$ and $(\mathbb{R}^{I_1 \times \cdots \times I_d}, \|\cdot\|_\sigma)$. There is no tensor analogue of Eckart--Young--Mirsky result: Frobenius and operator optima do not coincide.

\paragraph{Spectral suboptimality on heavy-tailed spectra.} Fix a target mode $\ell$ and matrix rank $k$. Let $W_\ell := \mathcal{T}_{(\ell)}$. We compare the Eckart--Young optimum $\widehat{W}^{\mathrm{EY}}$ of rank $k$ with the unfolding $\widehat{W} := (\widehat{\mathcal{T}})_{(\ell)}$ of the tensor truncation. HOSVD and TT-SVD do not solve the Frobenius problem over $\mathcal{M}_\mathbf{r}$ exactly, but satisfy the quasi-optimality bound $\|\mathcal{T} - \widehat{\mathcal{T}}\|_F \leq (1 + \varepsilon)\, \min_{\mathcal{S} \in \mathcal{M}_\mathbf{r}} \|\mathcal{T} - \mathcal{S}\|_F$ with $\varepsilon \leq \sqrt{d-1} - 1$~\citep{oseledets2011tensor}. Combined with the mode-wise tail estimate and the reshape isometry, this gives $\|W_\ell - \widehat{W}\|_F^2 \leq d \sum_{i > k} \sigma_i^2$ under the working assumption of comparable mode spectra.

\begin{corollary}[Spectral-to-Frobenius gap]
\label{cor:spectral-gap}
Combining $\|\cdot\|_2 \leq \|\cdot\|_F$ with the bound above,
\begin{equation}
\frac{\|W_\ell - \widehat{W}\|_2}{\sigma_{k+1}} \;\leq\; \sqrt{d} \cdot \sqrt{1 + \!\!\sum_{i > k+1}\! \bigl(\sigma_i / \sigma_{k+1}\bigr)^2}.
\label{eq:spectral-suboptimality}
\end{equation}
The manifold geometry contributes only $\sqrt{d}$; the binding factor is the spectral profile of $W_\ell$.
\end{corollary}

\paragraph{Power-law specialisation.}
Weight matrix spectra are well modelled by $\sigma_i \propto i^{-\alpha}$ over a broad index range~\citep{martin2021implicit}. Aggregating up to dimension $N$,
\begin{equation}
\sum_{i = k+1}^{N} \sigma_i^2 \;\asymp\; \begin{cases}
N^{1 - 2\alpha}, & \alpha < 1/2, \\
\log(N/k), & \alpha = 1/2, \\
k^{1 - 2\alpha}, & \alpha > 1/2.
\end{cases}
\label{eq:power-law-tail}
\end{equation}
Substituting into~\eqref{eq:spectral-suboptimality}, the spectral suboptimality grows as $\Theta(N^{1/2 - \alpha})$ for $\alpha < 1/2$, as $\Theta(\sqrt{\log(N/k)})$ at the boundary, and is bounded by a constant for $\alpha > 1/2$. The tail is dominated by the \emph{ambient dimension} $N$, not the truncation cutoff $k$, precisely in the heavy-tailed regime. The bound is consequential only when $\alpha < 1/2$. To verify that this is the operative regime, we fit $\sigma_i \propto i^{-\alpha_k}$ on the mode unfoldings of the stacked expert FFN tensors in Qwen3-30B-A3B (mode 0: expert; mode 1: output; mode 2: input). Table~\ref{tab:alpha} reports the exponents across representative layers: \emph{all measured $\alpha_k$ lie well below $1/2$}, with the expert and output modes routinely below $0.1$ and the input mode below $0.4$. The bound~\eqref{eq:spectral-suboptimality} is therefore not asymptotic ornamentation, but a practical binding at the dimensions and ranks used in practice.

\begin{table}[t]
\centering
\small
\caption{Fitted power-law exponents $\alpha_k$ ($\sigma_i \propto i^{-\alpha_k}$) for the three mode unfoldings of the expert tensors in Qwen3-30B-A3B.}
\label{tab:alpha}
\begin{tabular}{lccc}
\toprule
Layer / projection & $\alpha_0$ (expert) & $\alpha_1$ (output) & $\alpha_2$ (input) \\
\midrule
Layer 0 / gate  & 0.160 & 0.075 & 0.402 \\
Layer 0 / up    & 0.187 & 0.062 & 0.358 \\
Layer 12 / gate & 0.047 & 0.050 & 0.197 \\
Layer 12 / up   & 0.040 & 0.045 & 0.190 \\
Layer 24 / gate & 0.039 & 0.053 & 0.202 \\
Layer 24 / up   & 0.025 & 0.046 & 0.195 \\
Layer 47 / gate & 0.052 & 0.051 & 0.142 \\
Layer 47 / up   & 0.048 & 0.051 & 0.134 \\
\bottomrule
\end{tabular}
\vspace{-2mm}
\end{table}

\paragraph{Reconciliation with the empirical findings.} Proposition~\ref{prop:gap} and Corollary~\ref{cor:spectral-gap} together imply that any Frobenius-optimal tensorization controls $\|W - \widehat{W}\|_F$ while admitting an operator residual $\|W - \widehat{W}\|_2$ up to $\sqrt{\min(m,n)}$ times larger than the matrix-optimal value. Since residual-stream propagation is governed by operator rather than Frobenius norms, Frobenius optimality is misaligned with the quantity that determines functional preservation.

This account reconciles the findings of Section~\ref{sec:compression}. The angular drift and norm shrinkage in Fig.~\ref{fig:activation_geometry} track operator error, whereas the recovery curves of Fig.~\ref{fig:superweigths} reflect that the critical singular directions sit deep in the spectrum that Frobenius truncation discards first. The \textsc{TD-MoE} vs \textsc{MoBE} gap in Fig.~\ref{fig:moe_main} is the same effect amplified: tensorizing over the expert axis introduces a mode whose unfolding is the most heavy-tailed of all (Table~\ref{tab:alpha}, $\alpha_0 \approx 0.03\text{--}0.05$ at intermediate layers), multiplying an already strict bound. None of these failures can be mitigated by reweighting the Frobenius loss or reordering modes; they require either an operator-aware objective or post-decomposition repair substantially heavier than the lightweight LoRA used here.

\section{Comparison with the Quantization Baseline}
\label{sec:ptq-comparison}

Sections~\ref{sec:compression} and~\ref{sec:theory} established that tensor decompositions face a structural barrier as standalone post-training compressors. To place these findings in the broader landscape of training-free compression, we benchmark all matrix- and tensor-based methods of Section~\ref{sec:compression} against a standard post-training quantization: round-to-nearest (\texttt{RTN}) at 4 and 8 bits, applied to same Transformer layers and modules \citep{gholami2022survey}. Figure~\ref{fig:ptq-comparison} reports the resulting C4 perplexity versus bits saved on GPT-J~6B, including a new combination not considered earlier: Tucker on MHA paired with \texttt{LASER} (at ranks $512$ and $1024$) on FFN, which probes the best hybrid achievable from the methods surveyed in Section~\ref{sec:compression}. Additional experiments across different models and setups are presented in Appendix~\ref{app:quality_compression} and~\ref{app:td_moe_gpt_oss}.

\begin{wrapfigure}{r}{0.6\linewidth}
\vspace{-4mm}
\centering
\includegraphics[width=0.98\linewidth]{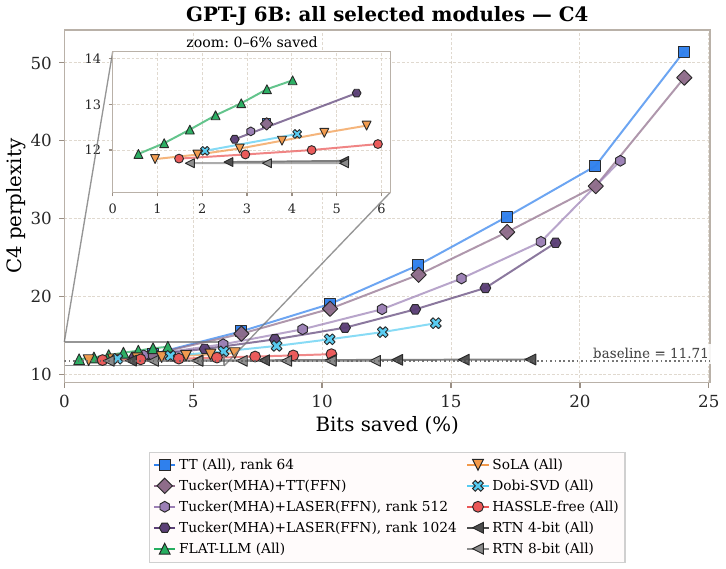}
\caption{C4 perplexity versus bits saved on GPT-J 6B for all selected modules and post-training quantization.}
\label{fig:ptq-comparison}
\vspace{-6mm}
\end{wrapfigure}

According to Figure~\ref{fig:ptq-comparison} \texttt{RTN} dominates the compression-quality frontier. Both \texttt{RTN}-4-bit and \texttt{RTN}-8-bit track the dense baseline across the full tested range, with no perceptible perplexity degradation up to $\approx 18\%$ bits saved. Among decomposition-based methods, only \texttt{HASSLE-free} remains genuinely competitive in the low-compression regime, where its structure absorbs the high-magnitude entries that pure low-rank truncation distorts; \texttt{SoLA} and \texttt{Dobi-SVD} follow at a slightly larger margin from the baseline. Beyond this regime, every matrix- and tensor-based method incurs perplexity penalties that \texttt{RTN} avoids by construction. This confirms that, at deployment-relevant compression ratios, training-free decompositions are not a substitute for quantization but at best a complementary tool.

At matched compression ratios in the mid range ($\approx 15\text{--}20\%$ bits saved), the hybrid Tucker(MHA)+\texttt{LASER}(FFN) variants outperform both Tucker(MHA)+TT(FFN) and TT(All) by a substantial margin: at $\approx 17\%$ bits saved, Tucker(MHA)+\texttt{LASER}(FFN) at rank $1024$ achieves $\mathrm{PPL}\approx 19$, whereas the TT-based combinations exceed $\mathrm{PPL}\approx 28$. Taken together, these results delineate the viable role of tensor decompositions in post-training LLM compression. They cannot compete with quantization as a standalone strategy, and within the family of decomposition methods they are dominated by matrix-level alternatives due to heavy-tailed spectra of FFN as Table~\ref{tab:alpha} shows.

\section{Conclusion}
This study shows that tensor decompositions fail as a standalone post-training LLM compression method because they optimize the wrong geometry. Tucker and TT preserve Frobenius mass while distorting the spectral (head-, neuron-, and expert-specific) subspaces that carry model function. These perturbations accumulate across depth, and manifest as activation drift and downstream degradation.

\section{Limitations}

The evaluation covers two dense architectures (GPT-J~6B and LLaMA~2~7B) and two MoE architectures (Qwen3-30B-A3B and GPT-OSS-20B). While these span a range of scales and designs, the conclusions may not transfer without verification to architectures that differ substantially in attention structure, normalisation, or expert routing, such as models trained with Muon or other second-order optimisers, which are known to produce qualitatively different weight spectra.

The theoretical analysis establishes operator-norm and spectral-suboptimality bounds that are sufficient conditions for the observed failure, but not tight characterisations. In particular, Proposition~\ref{prop:gap} and Corollary~\ref{cor:spectral-gap} describe worst-case separation between Frobenius- and operator-optimal truncation; the practical gap for a specific architecture may be smaller or larger depending on the singular-value profile of individual layers. The power-law exponents in Table~\ref{tab:alpha} are fitted to MoE expert tensors and may not generalise to all dense layer types.

The LoRA repair stage is intentionally lightweight (rank~16, 100
optimiser steps) and is intended as a lower bound on recoverability
rather than a best-effort fine-tuning baseline. Stronger post-hoc
adaptation could narrow the quality gap reported here; the paper makes no claims about the limits of recovery under full fine-tuning.

All models and benchmarks used in this work are publicly available and employed solely for research evaluation of compression methods, in accordance with their stated intended use. We do not redistribute model weights or derive products outside research contexts. The compression techniques studied here do not introduce new capabilities that raise safety concerns beyond those already present in the evaluated models; in particular, we study only post-training weight compression and do not fine-tune models on sensitive data or produce outputs intended for deployment.

\end{mainpart}

\newpage
\begin{appendixpart}
\vspace{-4mm}
\tableofcontents
\allowdisplaybreaks
\newpage

\section{Notation}
\label{app:notation}

\paragraph{General objects and conventions.}
Scalars are denoted by lowercase letters such as $r$, $p$, $q$, $\rho$, and $L$.
Matrices are denoted by uppercase letters such as $X$, $\widehat{X}$, $A$, $W$, $\widehat{W}$, $U$, $V$, and $\Sigma$. Tensors are denoted by calligraphic letters such as $\mathcal{X}$, $\mathcal{G}$, and $\mathcal{T}$. The notation $\widehat{\cdot}$ denotes an approximation of the corresponding dense object. The symbol $\approx$ denotes an approximate factorization or reconstruction.

\paragraph{Compression and evaluation variables.}
$L$ denotes the number of consecutive compressed Transformer blocks in the quality--compression trade-off experiments. The LoRA repair rank is denoted $r=16$. The TD-MoE per-layer compression ratio is denoted by $\rho$, with experiments using $\rho \in \{0.2,0.4\}$. In the GPT-OSS-20B expert-mode comparison, \textsc{preserve} fixes $r_1=K$, while \textsc{compress} uses $r_1<K$. Example selected TD-MoE ranks are $(32,1720,2664)$ for \textsc{preserve} and $(20,2496,2880)$ for \textsc{compress}.

\paragraph{Method abbreviations.}
Table~\ref{tab:method_abbreviations} lists the method-specific abbreviations used throughout the paper.

\begin{table}[htbp]
\centering
\caption{Method abbreviations.}
\label{tab:method_abbreviations}
\small
\begin{tabular}{@{}p{\columnwidth}@{}}
\toprule
\textbf{Abbreviation and meaning / role in the paper} \\
\midrule
\textbf{HOSVD} --- Higher-order singular value decomposition \\
\textbf{TT-SVD} --- SVD-based algorithm for constructing Tensor Train decompositions \\
\textbf{TD-MoE} --- Tensor-decomposition compression of Mixture-of-Experts layers \\
\textbf{LASER} --- Truncated-SVD-based post-training compression baseline for FFN compression \\
\textbf{LeSTD} --- Prior Tucker-style tensor-compression method for Transformer attention projections \\
\textbf{TensorLLM} --- Prior Tucker-style tensor-compression method for Transformer attention projections \\
\bottomrule
\end{tabular}
\end{table}





\section{Additional information and experiments for Dense models}
\subsection{GPT-J and LLaMA 2 Tensor-Decomposition Protocol}
\label{app:gptj_llama_tensor_protocol}

This section describes the post-training tensor-decomposition experiments on GPT-J 6B~\citep{gpt-j} and LLaMA 2 7B~\citep{touvron2023llama} reported in Figure~\ref{fig:c4_pareto}. We use the progressive block schedule: GPT-J 6B is compressed from block 14 to block 20, and LLaMA 2 7B from block 16 to block 23, adding one consecutive block at a time. All runs are evaluated relative to a fixed dense baseline for the same model.

For attention compression, we use a TensorLLM-style Tucker factorization of the query, key, value, and output projections. Since these projections have matching shapes in both models, we stack them into a tensor with input-feature, head, head-dimension, and projection-type modes. We apply partial Tucker factorization to the input-feature, head-dimension, and projection-type modes, leaving the head mode explicit. The maximum input-feature rank is 64, the head-dimension rank is 4, and the projection-type rank is 2.

For TT compression of FFN projections, and for the TT-all setting, each linear weight matrix is tensorized into twelve paired input-output modes. The input and output dimensions are split into twelve approximately balanced factors, interleaved into input-output pairs, and compressed with TT ranks capped at 64. Bias terms, when present, are kept dense.

LoRA repair is applied after decomposition with the decomposed modules frozen. We train rank-16 LoRA adapters with scaling factor 32 on WikiText-2 using AdamW, learning rate $2 \times 10^{-4}$, 100 optimizer steps, and gradient accumulation over 8 micro-batches. This stage tests whether limited data-dependent adaptation can recover quality lost by factorization, rather than performing full fine-tuning.

\begin{figure*}[ht]
  \centering
  \includegraphics[width=0.95\linewidth]{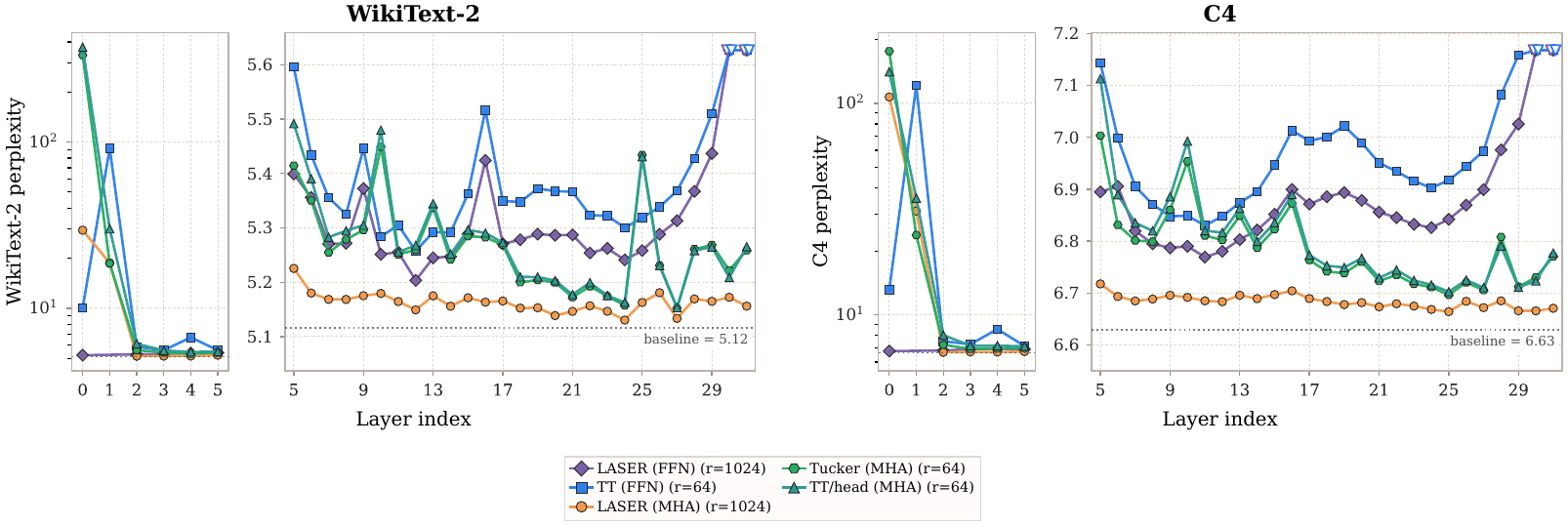}
  \caption{Single-layer decomposition sensitivity on LLaMA 2 7B. Each point compresses only one transformer layer. The left and right panels report WikiText-2 and C4 perplexity, respectively; early layers are shown separately because of large perplexity spikes.}
  \label{fig:app_single_layer_decomposition}
\end{figure*}

Perplexity is evaluated on WikiText-2 and C4, using sequence length 2048 for GPT-J 6B and 4096 for LLaMA 2 7B. For activation geometry diagnostics, we collect WikiText-2 sequences from the dense and compressed models and compare activations at the last compressed block. We report the mean angular deviation from the dense activations and the compressed-to-dense activation norm ratio. Storage is always computed from the compressed representation, excluding embeddings; for benchmarking, compressed modules are reconstructed to dense linear layers so that the reported quality reflects the compressed weights rather than implementation-specific runtime kernels.

\subsection{Single-Layer Decomposition on LLaMA 2 7B}
\label{app:single_layer_decomposition}

To separate local layer sensitivity from error accumulation across depth, we also evaluate a single-layer protocol on LLaMA 2 7B. In each run, only one transformer layer is modified, and all other layers remain dense. We repeat this for all 32 layers and evaluate WikiText-2 and C4 perplexity. For attention layers, we compare LASER-style matrix factorization, TensorLLM-style Tucker factorization, and a per-head TT variant. For FFN layers, we compare LASER-style matrix factorization with TT factorization. Tucker always refers to the TensorLLM-style decomposition of the attention projections.

Figure~\ref{fig:app_single_layer_decomposition} shows that sensitivity is highly non-uniform across depth. The earliest layers are especially fragile: decomposing the first attention layer causes very large perplexity spikes for Tucker and per-head TT, while TT on FFN is most unstable in the first two FFN layers. Away from these early layers, attention decompositions usually stay much closer to the dense baseline, although they save little of the total model size. FFN decompositions save more parameters, but their effect grows toward the final layers, especially for TT. This supports the progressive middle-to-late compression schedule used in the main experiments: middle layers are less sensitive locally, but quality still degrades once errors are accumulated over multiple compressed blocks.

It is also worth noting separately that the PPL of early layers,
particularly those close to the layer containing the super-weight
(Layer 1), degrades far more severely than even under the pruning
strategy(see Figure~\ref{fig:full_width_graph}). This further confirms that retaining only the dominant
singular components is insufficient to recover the structure of the
original weight matrix.

\begin{table}[ht]
\centering
\small
\setlength{\tabcolsep}{3.5pt}
\caption{Storage accounting for the LLaMA 2 7B single-layer decomposition study. Each run compresses one layer at a time. Bits saved are measured over non-embedding model parameters.}
\label{tab:app_single_layer_storage}
\begin{tabular}{@{}llccc@{}}
\toprule
\textbf{Method} & \textbf{Target} & \textbf{Rank} & \textbf{Local CR} & \textbf{Saved (\%)} \\
\midrule
LASER  & MHA & 1024 & 2.00   & 0.51 \\
Tucker & MHA & 64   & 240.49 & 1.01 \\
TT/head& MHA & 64   & 1.98   & 0.50 \\
LASER  & FFN & 1024 & 2.92   & 1.34 \\
TT     & FFN & 64   & 398.63 & 2.04 \\
\bottomrule
\end{tabular}
\end{table}

\subsection{Outlier Restoration Without Known Super-Weights}
\label{app:no-superweights}

The main restoration experiment in Section~\ref{sec:superweights} focuses on
the layer-1 MLP output projection of LLaMA~2~7B, where a known super-weight is
present. To test whether the observed rank-restoration barrier is specific to
that single coordinate, we repeat the same diagnostic on the corresponding MLP
output projection in layer~17. We track random inliers, top-magnitude entries, and
activation-weighted top-$k$ entries selected by
$|W_{ij}|\sqrt{\operatorname{diag}(X^\top X)_j}$.

\begin{figure*}[t]
  \centering
  \includegraphics[width=0.9\textwidth]{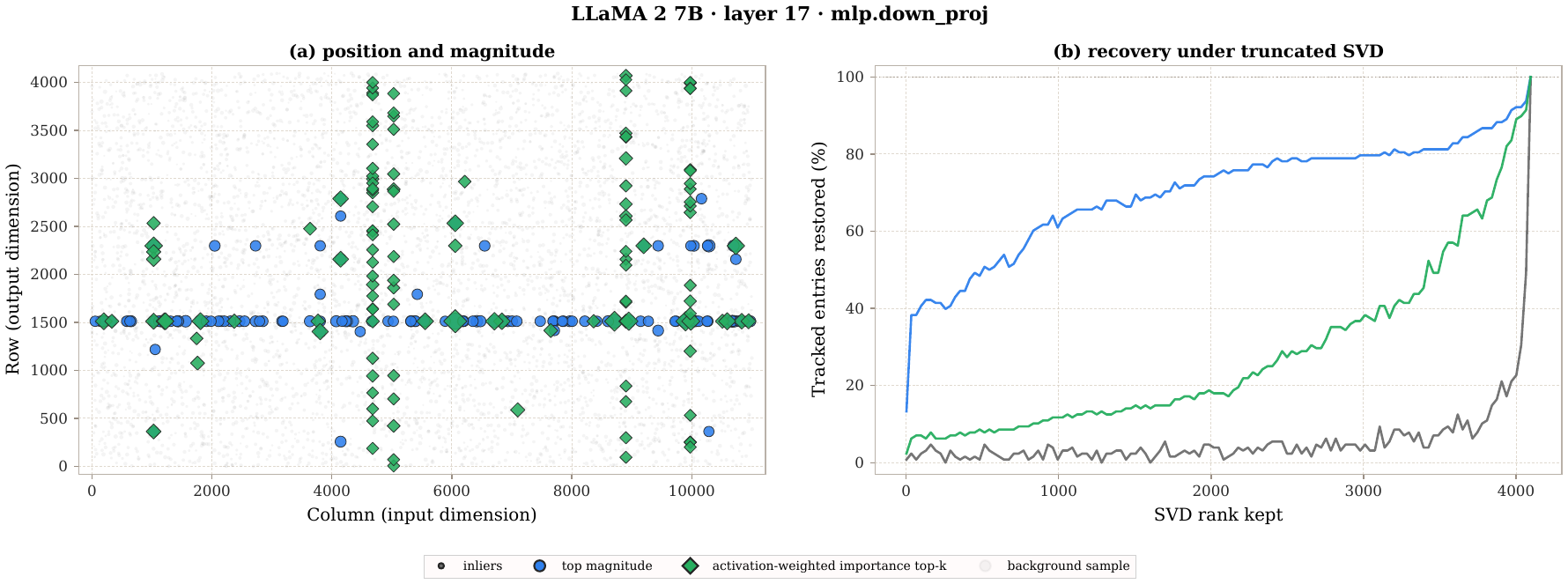}
  \caption{
    \textbf{(a)} Positions of random inliers, top-magnitude entries, and
    activation-weighted top-$k$ entries in the weight matrix. Marker size
    encodes absolute weight magnitude.
    \textbf{(b)} Fraction of tracked entries restored as a function of retained
    SVD rank.
  }
  \label{fig:app_layer17_outlier_restoration}
\end{figure*}

Figure~\ref{fig:app_layer17_outlier_restoration} shows that the qualitative
pattern from Figure~\ref{fig:superweigths} persists even without a
super-weight. Top-magnitude entries are recovered more smoothly as the rank
increases, but activation-weighted entries remain substantially below full
recovery until the retained rank approaches the full matrix rank.

\subsection{Detailed GPT-J and LLaMA 2 Results}
\label{app:detailed_gptj_llama_results}

Table~\ref{tab:app_main_tensor_results_full} reports the GPT-J 6B and LLaMA 2 7B tensor-decomposition experiments. Each cell is shown as \emph{direct / +LoRA}, where +LoRA denotes the lightweight rank-16 LoRA trained on WikiText-2. The maximum-$L$ table gives the most compressed setting for each model, while the full table reports all middle-to-late block ranges. For activation geometry, \emph{Angle} is the mean angle between dense and compressed residual-stream activations after the last compressed block, and \emph{Norm} is the compressed-to-dense activation norm ratio. Lower Angle and Norm closer to $1$ are better.
\subsection{Quality--Compression Trade-offs}
\label{app:quality_compression}

In addition to C4 perplexity, we evaluate WikiText-2 perplexity (Figure~\ref{fig:app_wt2_pareto}) and zero-shot LM-Eval accuracy. The LM-Eval score is computed on ARC-Challenge, HellaSwag, OpenBookQA, PIQA, and WinoGrande (Figure~\ref{fig:app_lmeval_pareto}). For each task, we measure the accuracy drop relative to the dense model in percentage points, and report the unweighted average across tasks as the macro LM-Eval accuracy drop. Lower values are better for all metrics shown in this section.

\begin{figure*}[ht]
  \centering
  \includegraphics[width=0.95\linewidth]{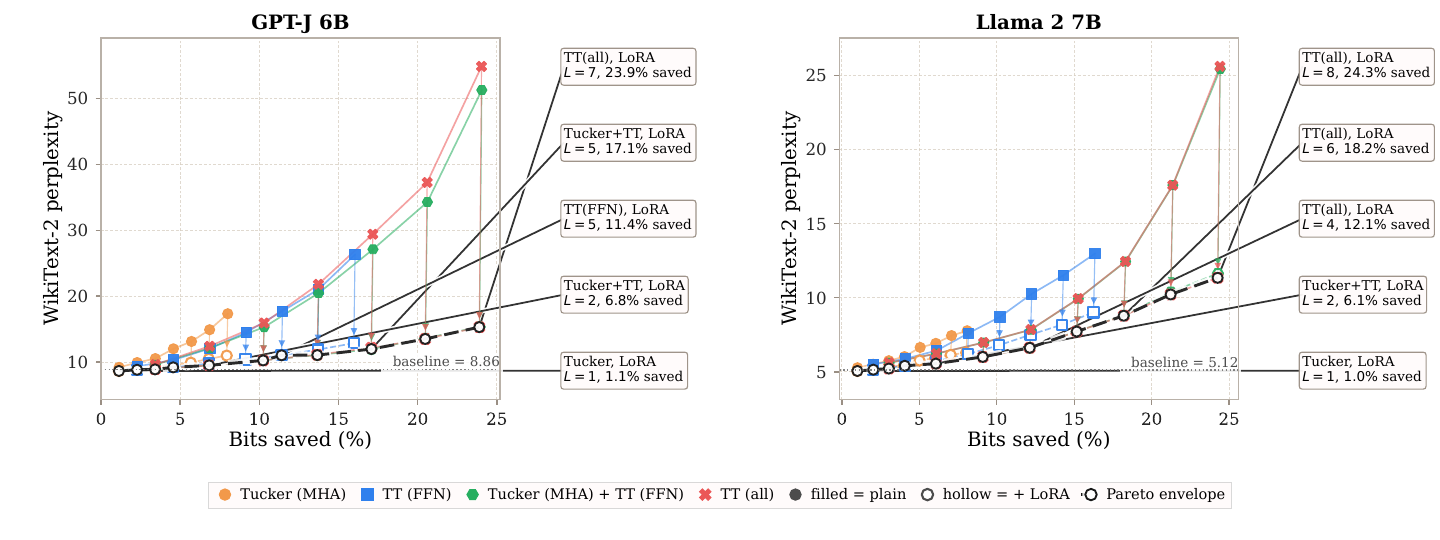}
  \caption{WikiText-2 perplexity versus bits saved for GPT-J 6B and LLaMA 2 7B. Each point is one compression run, and $L$ denotes the number of consecutive compressed transformer blocks. Arrows connect each decomposition to its LoRA-repaired variant; the Pareto frontier marks the best observed trade-offs.}
  \label{fig:app_wt2_pareto}
\end{figure*}

\begin{figure*}[ht]
  \centering
  \vspace{-4mm}\includegraphics[width=0.95\linewidth]{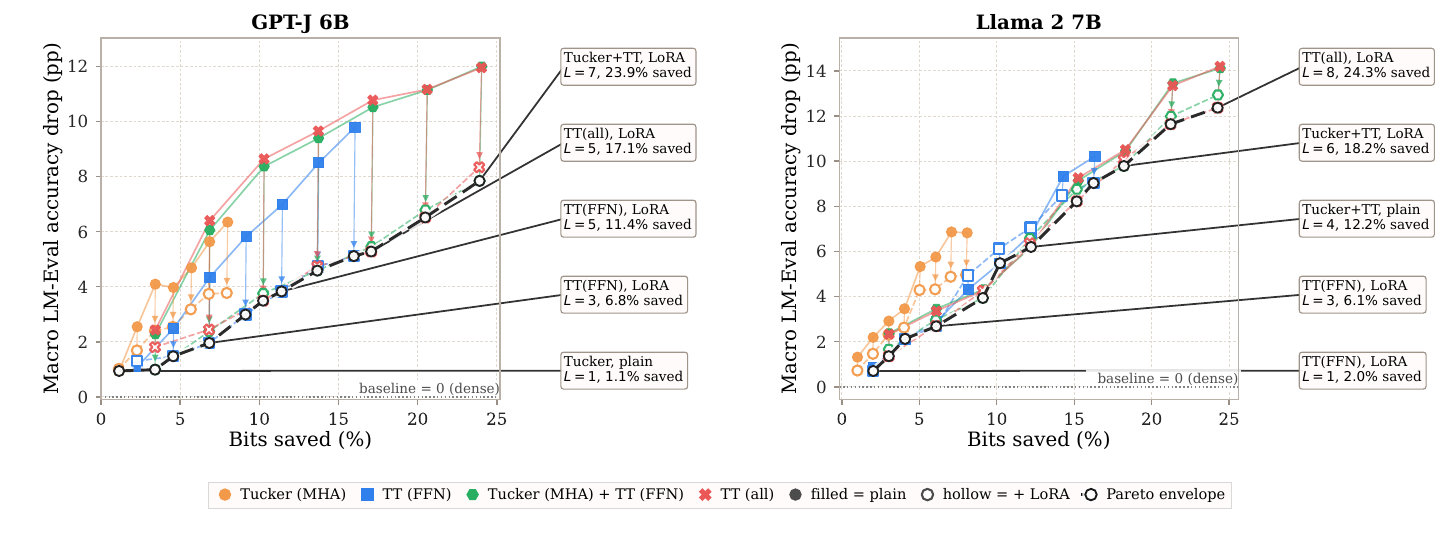}
  \caption{Macro LM-Eval accuracy drop versus bits saved for GPT-J 6B and LLaMA 2 7B. Macro drop is the unweighted average accuracy drop, in percentage points, across ARC-Challenge, HellaSwag, OpenBookQA, PIQA, and WinoGrande. Lower is better.}
  \label{fig:app_lmeval_pareto}
\end{figure*}

\subsection{Dense-Sparse TT with Outlier Preservation}
\label{app:dense_sparse_tt}

We also test whether explicitly preserving a small number of large or activation-important entries can mitigate TT degradation. For a target weight matrix $W$, Dense-Sparse TT first selects a sparse set $\mathcal{S}$ of entries, stores those entries exactly, and applies TT factorization only to the remaining inlier matrix:
\[
    W = W_{\mathcal{S}} + W_{\mathrm{in}},
    \qquad
    W_{\mathrm{in}} \approx \widehat{W}_{\mathrm{TT}} .
\]
The final compressed layer is therefore implemented as a TT layer plus a sparse correction. We evaluate two sparse-entry selectors. The magnitude-selected version chooses the top fraction $f$ of entries by $|W_{ij}|$. The activation-weighted version instead ranks entries by
\[
    \mathrm{score}_{ij}
    =
    |W_{ij}|
    \sqrt{\operatorname{diag}(X^\top X)_j},
\]
where $X$ contains calibration inputs to the corresponding linear layer. This score is inspired by the activation-aware sparse selection used in HASSLE-free, but we use it only to choose the sparse correction entries; the remaining matrix is still compressed with the same TT procedure. We collect the activation statistics using 64 calibration sequences of length 256.

Figure~\ref{fig:app_dense_sparse_tt_all} shows that preserving a tiny sparse correction substantially improves GPT-J 6B relative to plain TT when all selected modules are compressed. At the largest compressed range, plain TT reaches much higher C4 perplexity, whereas both Dense-Sparse variants remain close to each other and substantially lower. On LLaMA 2 7B, however, the Dense-Sparse variants are almost indistinguishable from TT. Overall, sparse outlier preservation helps in some cases, especially GPT-J, but it does not remove the model-dependent compression--quality trade-off.

\clearpage
\begingroup
\tiny
\setlength{\tabcolsep}{0.5pt}
\begin{longtable}{@{}ccp{1.5cm}ccccccccccc@{}}
\caption{Detailed GPT-J 6B and LLaMA 2 7B tensor-decomposition results. Compressed rows are grouped by the number of consecutive decomposed blocks $L$ and report \emph{direct/+LoRA}. Bits saved are measured over all non-embedding model parameters. Bold values are best among compressed runs within each model-$L$ group.}\label{tab:app_main_tensor_results_full}\\
\toprule
 & & & \multicolumn{1}{c}{\textbf{Compression} $\uparrow$} & \multicolumn{2}{c}{\textbf{Perplexity} $\downarrow$} & \multicolumn{6}{c}{\textbf{LM-Eval accuracy (\%)} $\uparrow$} & \multicolumn{2}{c}{\textbf{Activation geometry}} \\
\cmidrule(lr){4-4}\cmidrule(lr){5-6}\cmidrule(lr){7-12}\cmidrule(lr){13-14}
\textbf{$L$} & \textbf{Blocks} & \textbf{Method} & \makecell{Bits saved\\(\%)} & \textbf{WT2} & \textbf{C4} & \textbf{Macro} & \textbf{ARC-C} & \textbf{HSwag} & \textbf{OBQA} & \textbf{PIQA} & \textbf{WinoG} & \makecell{Angle\\$\downarrow$} & \makecell{Norm\\$\to 1$} \\
\midrule
\endfirsthead
\caption[]{Detailed GPT-J 6B and LLaMA 2 7B tensor-decomposition results (continued).}\\
\toprule
 & & & \multicolumn{1}{c}{\textbf{Compression} $\uparrow$} & \multicolumn{2}{c}{\textbf{Perplexity} $\downarrow$} & \multicolumn{6}{c}{\textbf{LM-Eval accuracy (\%)} $\uparrow$} & \multicolumn{2}{c}{\textbf{Activation geometry}} \\
\cmidrule(lr){4-4}\cmidrule(lr){5-6}\cmidrule(lr){7-12}\cmidrule(lr){13-14}
\textbf{$L$} & \textbf{Blocks} & \textbf{Method} & \makecell{Bits saved\\(\%)} & \textbf{WT2} & \textbf{C4} & \textbf{Macro} & \textbf{ARC-C} & \textbf{HSwag} & \textbf{OBQA} & \textbf{PIQA} & \textbf{WinoG} & \makecell{Angle\\$\downarrow$} & \makecell{Norm\\$\to 1$} \\
\midrule
\endhead
\midrule
\multicolumn{14}{r}{\emph{Continued on next page}}\\
\endfoot
\bottomrule
\endlastfoot
\multicolumn{3}{@{}l}{\makecell[l]{\textbf{GPT-J 6B}\\Dense baseline}} & 0.0 & 8.86 & 11.71 & 50.5 & 33.9 & 49.5 & 29.0 & 75.5 & 64.4 & -- & -- \\
\midrule
\multirow{4}{*}{1} & \multirow{4}{*}{14} & Tucker MHA & 1.1/1.1 & 9.19/\textbf{8.65} & \textbf{12.01}/12.30 & \textbf{49.5}/49.4 & 32.6/\textbf{34.1} & 47.8/\textbf{49.2} & 28.0/25.8 & 74.6/\textbf{74.9} & \textbf{64.6}/63.1 & \textbf{9.3}/18.3 & \textbf{0.99}/1.04 \\
 &  & TT FFN & 2.3/2.3 & 9.32/8.83 & 12.35/17.33 & 49.4/49.2 & 32.5/33.4 & 47.7/49.1 & 27.8/27.2 & 74.5/73.6 & 64.2/62.6 & 14.9/23.5 & 0.99/1.05 \\
 &  & Tucker+TT & \textbf{3.4}/3.4 & 9.60/8.89 & 12.57/13.41 & 48.2/49.5 & 29.8/33.0 & 46.2/48.4 & 26.6/\textbf{29.4} & 74.3/73.9 & 64.1/62.6 & 16.9/24.6 & 0.98/1.03 \\
 &  & TT all & 3.4/3.4 & 9.62/8.93 & 12.59/13.74 & 48.0/48.7 & 29.7/31.7 & 46.2/48.8 & 26.4/26.2 & 74.4/73.9 & 63.5/62.7 & 16.9/24.7 & 0.98/1.02 \\
\specialrule{0.10pt}{1pt}{1pt}
\multirow{4}{*}{2} & \multirow{4}{*}{14--15} & Tucker MHA & 2.3/2.3 & 9.93/\textbf{8.93} & \textbf{12.57}/12.71 & 47.9/48.8 & 30.2/32.8 & 46.5/\textbf{48.1} & 25.6/25.0 & 73.7/\textbf{74.1} & 63.5/\textbf{63.9} & \textbf{11.9}/20.3 & 0.98/1.04 \\
 &  & TT FFN & 4.6/4.6 & 10.31/9.20 & 13.59/14.39 & 48.0/\textbf{49.0} & 31.1/\textbf{33.5} & 45.5/48.0 & \textbf{27.8}/27.6 & 73.2/73.5 & 62.3/62.3 & 18.5/23.3 & 0.97/1.01 \\
 &  & Tucker+TT & \textbf{6.9}/6.8 & 12.13/9.50 & 15.20/13.91 & 44.4/48.1 & 26.3/31.7 & 41.7/47.0 & 23.0/26.0 & 71.3/73.8 & 59.8/62.0 & 20.5/26.2 & 0.97/\textbf{0.99} \\
 &  & TT all & 6.9/6.8 & 12.41/9.50 & 15.50/13.82 & 44.1/48.0 & 25.6/32.6 & 41.5/46.7 & 23.0/24.2 & 70.8/73.7 & 59.4/63.0 & 20.5/25.9 & 0.97/0.99 \\
\specialrule{0.10pt}{1pt}{1pt}
\multirow{4}{*}{3} & \multirow{4}{*}{14--16} & Tucker MHA & 3.4/3.4 & 10.53/\textbf{9.26} & \textbf{13.28}/13.30 & 46.4/48.1 & 28.5/31.8 & 44.9/\textbf{47.2} & 24.4/25.4 & 72.6/\textbf{73.3} & 61.4/\textbf{62.6} & \textbf{13.7}/19.9 & 0.98/\textbf{1.01} \\
 &  & TT FFN & 6.9/6.8 & 12.06/9.77 & 15.54/16.12 & 46.1/\textbf{48.5} & 28.4/\textbf{32.7} & 43.2/46.9 & 25.4/\textbf{28.4} & 72.9/73.2 & 60.7/61.4 & 21.5/24.7 & 0.94/0.97 \\
 &  & Tucker+TT & \textbf{10.3}/10.2 & 15.30/10.27 & 18.41/15.01 & 42.1/46.7 & 23.3/30.4 & 39.3/45.5 & 20.0/25.0 & 69.4/72.1 & 58.6/60.5 & 23.4/27.0 & 0.94/0.95 \\
 &  & TT all & 10.3/10.2 & 15.93/10.25 & 19.01/15.04 & 41.8/47.0 & 23.0/30.8 & 38.9/45.4 & 19.2/25.6 & 69.3/72.9 & 58.7/60.2 & 23.4/27.1 & 0.95/0.95 \\
\specialrule{0.10pt}{1pt}{1pt}
\multirow{4}{*}{4} & \multirow{4}{*}{14--17} & Tucker MHA & 4.6/4.5 & 12.01/\textbf{9.57} & 13.79/\textbf{13.61} & 46.5/\textbf{47.9} & 28.8/31.5 & 44.4/\textbf{47.0} & 24.8/25.2 & 72.9/\textbf{73.4} & 61.5/\textbf{62.5} & \textbf{15.2}/20.3 & 0.97/\textbf{1.00} \\
 &  & TT FFN & 9.2/9.1 & 14.49/10.38 & 18.28/16.99 & 44.6/47.5 & 26.7/\textbf{32.3} & 41.4/46.0 & 22.6/\textbf{26.4} & 71.5/72.8 & 61.0/59.9 & 24.1/26.1 & 0.92/0.94 \\
 &  & Tucker+TT & \textbf{13.7}/13.7 & 20.46/11.06 & 22.77/17.78 & 41.1/45.9 & 22.9/29.4 & 37.4/44.3 & 17.8/23.8 & 68.7/71.3 & 58.6/60.6 & 26.0/28.6 & 0.92/0.91 \\
 &  & TT all & 13.7/13.7 & 21.80/11.12 & 24.01/18.05 & 40.8/45.7 & 23.0/29.4 & 37.1/44.1 & 17.6/23.2 & 68.3/71.9 & 58.1/60.1 & 26.0/29.2 & 0.93/0.92 \\
\specialrule{0.10pt}{1pt}{1pt}
\multirow{4}{*}{5} & \multirow{4}{*}{14--18} & Tucker MHA & 5.7/5.7 & 13.12/\textbf{9.91} & 14.92/\textbf{14.20} & 45.8/\textbf{47.3} & 27.6/30.1 & 43.1/\textbf{46.2} & 24.6/\textbf{25.6} & 72.1/\textbf{73.1} & \textbf{61.5}/\textbf{61.5} & \textbf{16.6}/21.6 & 0.98/\textbf{1.00} \\
 &  & TT FFN & 11.5/11.4 & 17.73/11.03 & 21.46/18.36 & 43.5/46.6 & 27.0/\textbf{32.1} & 40.2/45.3 & 20.8/24.8 & 70.2/71.7 & 59.1/59.3 & 26.4/27.4 & 0.89/0.91 \\
 &  & Tucker+TT & \textbf{17.2}/17.1 & 27.10/11.97 & 28.26/19.00 & 39.9/45.0 & 22.3/29.7 & 35.9/43.2 & 17.8/22.6 & 65.8/70.9 & 57.9/58.6 & 28.1/30.2 & 0.90/0.89 \\
 &  & TT all & 17.2/17.1 & 29.39/12.16 & 30.22/20.33 & 39.7/45.2 & 22.1/29.4 & 35.6/43.1 & 17.4/23.4 & 65.7/71.0 & 57.6/59.0 & 28.2/31.2 & 0.91/0.89 \\
\specialrule{0.10pt}{1pt}{1pt}
\multirow{4}{*}{6} & \multirow{4}{*}{14--19} & Tucker MHA & 6.9/6.8 & 14.90/\textbf{10.48} & 16.04/\textbf{15.11} & 44.8/\textbf{46.7} & 26.8/\textbf{30.1} & 42.4/\textbf{45.9} & 22.4/23.2 & 71.2/\textbf{73.1} & \textbf{61.4}/61.3 & \textbf{17.9}/23.8 & 0.98/\textbf{0.99} \\
 &  & TT FFN & 13.8/13.7 & 21.05/11.89 & 24.54/20.10 & 42.0/45.7 & 24.7/28.8 & 38.6/43.9 & 20.2/\textbf{24.6} & 68.3/71.4 & 58.0/60.0 & 28.5/28.9 & 0.86/0.88 \\
 &  & Tucker+TT & \textbf{20.6}/20.5 & 34.27/13.62 & 34.16/22.35 & 39.3/43.7 & 21.6/26.8 & 34.6/41.3 & 18.0/21.0 & 64.4/69.9 & 58.0/59.4 & 30.1/33.3 & 0.87/0.86 \\
 &  & TT all & 20.6/20.5 & 37.25/13.50 & 36.73/21.31 & 39.3/43.9 & 22.0/26.2 & 34.5/41.4 & 17.6/22.6 & 64.3/70.8 & 58.1/58.8 & 30.1/33.0 & 0.88/0.86 \\
\specialrule{0.10pt}{1pt}{1pt}
\multirow{4}{*}{7} & \multirow{4}{*}{14--20} & Tucker MHA & 8.0/7.9 & 17.33/\textbf{10.98} & 17.56/\textbf{15.99} & 44.1/\textbf{46.7} & 26.5/\textbf{30.3} & 41.3/\textbf{45.4} & 21.0/\textbf{24.2} & 70.0/\textbf{71.7} & \textbf{61.9}/\textbf{61.9} & \textbf{19.4}/24.2 & \textbf{0.98}/0.98 \\
 &  & TT FFN & 16.0/16.0 & 26.31/12.86 & 28.28/20.00 & 40.7/45.4 & 24.1/29.2 & 37.3/43.4 & 17.8/23.8 & 66.2/71.1 & 57.9/59.3 & 30.5/30.3 & 0.83/0.87 \\
 &  & Tucker+TT & \textbf{24.1}/23.9 & 51.26/15.34 & 48.09/21.51 & 38.5/42.6 & 21.8/24.5 & 33.3/40.6 & 16.8/20.4 & 62.8/68.8 & 57.5/58.8 & 31.8/35.1 & 0.85/0.85 \\
 &  & TT all & 24.0/23.9 & 54.84/15.30 & 51.42/21.99 & 38.5/42.1 & 21.9/24.8 & 33.0/40.3 & 17.8/19.8 & 62.8/68.4 & 57.1/57.3 & 31.8/35.1 & 0.86/0.85 \\
\midrule
\multicolumn{3}{@{}l}{\makecell[l]{\textbf{LLaMA 2 7B}\\Dense baseline}} & 0.0 & 5.12 & 6.63 & 56.2 & 43.0 & 57.1 & 33.4 & 78.1 & 69.4 & -- & -- \\
\midrule
\multirow{4}{*}{1} & \multirow{4}{*}{16} & Tucker MHA & 1.0/1.0 & 5.28/\textbf{5.06} & \textbf{6.87}/6.91 & 54.9/55.5 & 41.0/\textbf{42.7} & 55.3/\textbf{56.0} & 32.4/33.2 & 76.8/76.9 & 69.0/68.5 & \textbf{17.1}/22.5 & 0.91/0.99 \\
 &  & TT FFN & 2.0/2.0 & 5.52/5.15 & 7.01/7.05 & 55.3/\textbf{55.5} & 41.9/42.2 & 55.9/55.9 & 31.4/32.4 & 77.6/\textbf{78.0} & \textbf{69.9}/69.0 & 22.3/26.3 & 0.93/\textbf{1.00} \\
 &  & Tucker+TT & \textbf{3.1}/3.0 & 5.61/5.24 & 7.23/7.25 & 53.8/54.5 & 38.9/40.2 & 54.3/54.8 & 30.4/32.6 & 76.4/77.0 & 69.1/68.1 & 26.5/31.1 & 0.85/0.92 \\
 &  & TT all & 3.1/3.0 & 5.61/5.23 & 7.23/7.26 & 53.9/54.8 & 39.1/40.7 & 54.4/54.8 & 30.4/\textbf{33.6} & 76.4/77.1 & 69.1/68.0 & 26.5/31.5 & 0.85/0.92 \\
\specialrule{0.10pt}{1pt}{1pt}
\multirow{4}{*}{2} & \multirow{4}{*}{16--17} & Tucker MHA & 2.0/2.0 & 5.52/\textbf{5.20} & \textbf{7.15}/7.18 & 54.0/\textbf{54.7} & 38.8/\textbf{40.1} & 54.1/\textbf{55.2} & 31.6/\textbf{32.8} & 76.6/\textbf{77.0} & \textbf{69.0}/68.6 & \textbf{19.7}/23.9 & 0.89/\textbf{0.96} \\
 &  & TT FFN & 4.1/4.1 & 5.88/5.41 & 7.56/7.56 & 54.1/54.1 & 39.9/39.1 & 54.4/54.6 & 31.2/31.4 & 76.6/76.6 & 68.3/68.7 & 26.2/28.7 & 0.90/0.94 \\
 &  & Tucker+TT & \textbf{6.1}/6.1 & 6.24/5.56 & 7.97/7.88 & 52.7/53.2 & 37.8/38.0 & 52.2/53.1 & 29.8/30.4 & 75.2/76.3 & 68.7/68.4 & 30.5/34.1 & 0.80/0.86 \\
 &  & TT all & 6.1/6.1 & 6.24/5.56 & 7.97/7.90 & 52.8/53.5 & 38.0/38.7 & 52.2/53.2 & 29.8/31.4 & 75.2/76.3 & \textbf{69.0}/68.0 & 30.5/34.5 & 0.80/0.87 \\
\specialrule{0.10pt}{1pt}{1pt}
\multirow{4}{*}{3} & \multirow{4}{*}{16--18} & Tucker MHA & 3.0/3.0 & 5.76/\textbf{5.34} & 7.50/\textbf{7.45} & 53.3/\textbf{53.9} & 37.7/38.8 & 53.3/\textbf{54.6} & 30.8/\textbf{31.4} & 76.3/\textbf{76.6} & 68.4/68.0 & \textbf{21.7}/25.8 & 0.87/\textbf{0.96} \\
 &  & TT FFN & 6.1/6.1 & 6.44/5.73 & 8.33/8.17 & 53.3/53.5 & \textbf{39.4}/38.9 & 52.8/53.3 & 29.6/\textbf{31.4} & 76.0/75.7 & \textbf{68.7}/68.3 & 29.1/31.8 & 0.87/0.90 \\
 &  & Tucker+TT & \textbf{9.2}/9.1 & 6.98/6.03 & 8.95/8.66 & 51.9/52.3 & 35.5/36.5 & 50.5/51.4 & 30.8/30.0 & 74.7/76.0 & 67.9/67.5 & 33.5/36.7 & 0.76/0.81 \\
 &  & TT all & 9.2/9.1 & 6.98/6.01 & 8.95/8.65 & 51.9/51.9 & 35.5/36.1 & 50.5/51.3 & 30.8/30.0 & 74.8/75.0 & 68.1/67.2 & 33.6/36.8 & 0.76/0.82 \\
\specialrule{0.10pt}{1pt}{1pt}
\multirow{4}{*}{4} & \multirow{4}{*}{16--19} & Tucker MHA & 4.0/4.0 & 6.03/\textbf{5.46} & 7.91/\textbf{7.68} & 52.7/\textbf{53.6} & 36.9/\textbf{38.1} & 52.6/\textbf{54.0} & 30.2/\textbf{32.0} & 76.0/\textbf{76.2} & \textbf{68.0}/67.6 & \textbf{23.5}/25.8 & 0.86/\textbf{0.94} \\
 &  & TT FFN & 8.2/8.1 & 7.61/6.20 & 9.59/8.90 & 51.9/51.3 & 37.0/35.9 & 51.1/51.7 & 29.0/26.8 & 74.9/74.1 & 67.3/67.8 & 31.5/35.0 & 0.84/0.89 \\
 &  & Tucker+TT & \textbf{12.2}/12.1 & 7.85/6.63 & 10.06/9.62 & 50.0/49.6 & 35.0/34.4 & 49.0/49.4 & 25.6/24.6 & 73.6/73.8 & 66.9/66.0 & 36.0/39.3 & 0.72/0.77 \\
 &  & TT all & 12.2/12.1 & 7.86/6.63 & 10.06/9.65 & 50.0/49.8 & 34.9/34.1 & 49.0/49.5 & 25.2/25.6 & 73.6/73.6 & 67.0/66.4 & 36.0/38.9 & 0.72/0.78 \\
\specialrule{0.10pt}{1pt}{1pt}
\multirow{4}{*}{5} & \multirow{4}{*}{16--20} & Tucker MHA & 5.1/5.0 & 6.65/\textbf{5.77} & 8.64/\textbf{8.25} & 50.9/\textbf{51.9} & 34.5/36.4 & 51.0/\textbf{53.0} & 26.0/\textbf{27.6} & 74.1/\textbf{74.6} & \textbf{68.8}/68.0 & \textbf{25.3}/27.2 & 0.84/\textbf{0.92} \\
 &  & TT FFN & 10.2/10.2 & 8.71/6.80 & 10.99/9.73 & 50.7/50.1 & \textbf{36.5}/33.9 & 49.2/50.4 & \textbf{27.6}/26.2 & 73.6/73.0 & 66.8/66.9 & 33.5/36.1 & 0.81/0.84 \\
 &  & Tucker+TT & \textbf{15.3}/15.2 & 9.94/7.74 & 12.34/11.09 & 47.1/47.4 & 30.6/30.8 & 45.2/46.8 & 23.2/23.2 & 70.0/71.0 & 66.3/65.4 & 38.3/42.3 & 0.68/0.74 \\
 &  & TT all & 15.3/15.2 & 9.94/7.72 & 12.35/11.11 & 46.9/48.0 & 30.5/31.1 & 45.2/46.8 & 22.8/24.0 & 70.0/70.9 & 66.2/67.1 & 38.4/41.5 & 0.68/0.74 \\
\specialrule{0.10pt}{1pt}{1pt}
\multirow{4}{*}{6} & \multirow{4}{*}{16--21} & Tucker MHA & 6.1/6.0 & 6.93/\textbf{5.94} & 9.07/\textbf{8.50} & 50.4/\textbf{51.9} & 34.4/\textbf{36.3} & 50.3/\textbf{52.2} & 25.8/\textbf{28.2} & 73.2/\textbf{74.1} & 68.5/\textbf{68.6} & \textbf{26.7}/28.1 & 0.83/\textbf{0.91} \\
 &  & TT FFN & 12.3/12.2 & 10.26/7.52 & 12.51/10.98 & 49.4/49.1 & 35.6/32.3 & 47.2/48.8 & 26.2/25.6 & 71.3/72.0 & 66.9/66.9 & 35.2/38.3 & 0.79/0.81 \\
 &  & Tucker+TT & \textbf{18.3}/18.2 & 12.45/8.81 & 14.96/12.77 & 45.7/46.4 & 29.8/28.7 & 43.0/45.2 & 23.0/24.4 & 67.6/68.6 & 65.3/65.2 & 40.4/44.5 & 0.64/0.71 \\
 &  & TT all & 18.3/18.2 & 12.46/8.79 & 14.97/12.76 & 45.7/46.1 & 29.5/29.4 & 43.0/45.0 & 23.0/23.0 & 67.6/68.9 & 65.3/64.5 & 40.4/43.8 & 0.64/0.71 \\
\specialrule{0.10pt}{1pt}{1pt}
\multirow{4}{*}{7} & \multirow{4}{*}{16--22} & Tucker MHA & 7.1/7.0 & 7.45/\textbf{6.15} & 9.74/\textbf{8.97} & 49.3/\textbf{51.3} & 32.8/\textbf{35.6} & 49.8/\textbf{51.9} & 23.4/\textbf{27.2} & 72.0/\textbf{74.1} & \textbf{68.7}/67.8 & \textbf{28.1}/29.3 & 0.81/\textbf{0.91} \\
 &  & TT FFN & 14.3/14.2 & 11.52/8.19 & 14.14/12.24 & 46.9/47.7 & 32.0/31.1 & 45.1/47.5 & 23.0/23.8 & 69.3/70.3 & 65.0/65.9 & 36.7/40.7 & 0.76/0.80 \\
 &  & Tucker+TT & \textbf{21.4}/21.2 & 17.60/10.39 & 19.45/15.22 & 42.7/44.2 & 27.2/27.5 & 39.5/42.4 & 19.2/21.2 & 64.3/66.2 & 63.5/63.8 & 42.2/46.8 & 0.61/0.68 \\
 &  & TT all & 21.4/21.2 & 17.60/10.23 & 19.44/14.90 & 42.9/44.6 & 27.6/28.1 & 39.5/42.3 & 19.2/22.0 & 64.4/66.3 & 63.6/64.2 & 42.2/45.4 & 0.61/0.70 \\
\specialrule{0.10pt}{1pt}{1pt}
\multirow{4}{*}{8} & \multirow{4}{*}{16--23} & Tucker MHA & 8.1/8.0 & 7.78/\textbf{6.30} & 10.14/\textbf{9.36} & 49.4/\textbf{51.2} & 33.0/\textbf{35.8} & 49.5/\textbf{52.0} & 24.8/\textbf{26.8} & 71.6/\textbf{73.7} & \textbf{68.0}/\textbf{68.0} & \textbf{29.4}/29.9 & 0.80/\textbf{0.90} \\
 &  & TT FFN & 16.3/16.2 & 12.98/9.02 & 15.77/13.40 & 46.0/47.2 & 31.6/30.7 & 43.4/45.9 & 22.8/23.6 & 67.4/68.9 & 64.7/66.7 & 38.0/41.7 & 0.74/0.78 \\
 &  & Tucker+TT & \textbf{24.4}/24.3 & 25.44/11.64 & 23.53/17.15 & 42.1/43.3 & 27.8/27.7 & 37.7/40.9 & 18.8/18.8 & 62.6/64.6 & 63.5/64.3 & 43.8/48.7 & 0.58/0.65 \\
 &  & TT all & 24.4/24.3 & 25.61/11.35 & 23.59/16.61 & 42.0/43.8 & 27.7/27.8 & 37.7/41.1 & 18.4/20.6 & 62.6/64.5 & 63.5/65.2 & 43.9/47.1 & 0.58/0.68 \\
\end{longtable}
\endgroup

\begin{figure*}[ht]
  \centering
  \includegraphics[width=0.9\linewidth]{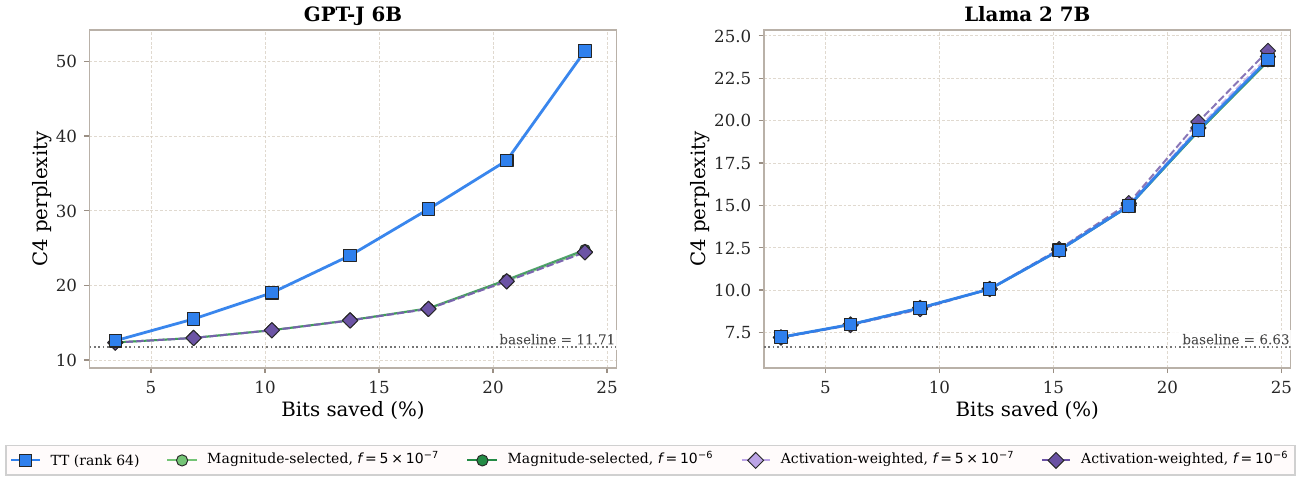}
  \caption{
    C4 perplexity versus bits saved for all selected modules under TT and Dense-Sparse TT. 
    Dense-Sparse TT stores a small sparse correction exactly and applies TT to the inlier matrix. 
    We compare magnitude-selected sparse entries with activation-weighted sparse entries for $f \in \{5{\times}10^{-7},10^{-6}\}$.
  }
  \label{fig:app_dense_sparse_tt_all}
\end{figure*}

\section{TD-MoE Experiments}
\label{sec:td_moe_qwen}

\subsection{Qwen3-30B-A3B and GPT-OSS-20B Tensor-Decomposition Protocol}
\label{app:td_moe}

This section describes the post-training tensor-decomposition experiments on Qwen3-30B-A3B~\citep{qwen3technicalreport} and GPT-OSS-20B~\citep{openai2025gptoss} reported in Figure~\ref{fig:moe_main}. We use the progressive block schedule. Qwen3-30B-A3B is compressed from the middle, the 24th layer. We first extend the compressed set toward later layers for 12 MoE layers, reaching roughly the 75\% depth point of the model, and then extend toward earlier layers for another 12 MoE layers, reaching roughly the 25\% depth point. The final setting therefore covers 24 MoE layers in total.
GPT-OSS-20B is compressed from the middle, the 12th layer. We first extend the compressed set toward later layers for 6 MoE layers, reaching roughly the 75\% depth point of the model, and then extend toward earlier layers for another 6 MoE layers, reaching roughly the 25\% depth point. The final setting therefore covers 12 MoE layers in total. We use two per-layer compression ratios $\rho \in \{0.2, 0.4\}$, where
$\rho$ controls how aggressively each expert is compressed via Tucker
decomposition (Figure~\ref{fig:qwen3_c4_ppl}). All runs are evaluated relative to a fixed dense baseline for the same model.

TD-MoE is a post-training compression method designed for Mixture-of-Experts
layers. Instead of decomposing each expert weight matrix independently, it
stacks all experts in a layer into a three-dimensional tensor over expert,
input, and output modes, then applies a joint Tucker factorization:
\[
\mathcal{X} \approx \mathcal{G} \times_1 U^{(1)} \times_2 U^{(2)} \times_3 U^{(3)} .
\]
Here $U^{(1)} \in \mathbb{R}^{K \times r_1}$ acts on the expert mode and
represents $r_1$ latent meta-experts, $U^{(2)} \in \mathbb{R}^{d_{\mathrm{in}}
\times r_2}$ spans the compressed input-feature subspace, and
$U^{(3)} \in \mathbb{R}^{d_{\mathrm{out}} \times r_3}$ spans the compressed
output-feature subspace. The core tensor $\mathcal{G}$ couples these three latent modes.

Figure~\ref{fig:td_moe_downstream} reports the corresponding downstream
accuracy curves. A plausible explanation for the model-dependent behavior is
that Qwen3-30B-A3B is a fine-grained MoE with 128 experts and 8 active experts
per token. Recent work identifies three shallow
\emph{super experts} in layers 1--3 whose pruning raises WikiText-2 perplexity
from 8.70 to 59.86 and collapses reasoning, while randomly pruning non-super
experts has negligible effect~\citep{su2025superexperts}. Our schedule starts at mid-depth and does not touch those shallow super experts, so
moderate TD-MoE may act mainly as denoising in less critical expert subspaces,
consistent with prior observations that selective rank reduction can sometimes
improve LLM accuracy by removing harmful higher-order components~\citep{LASER_2024}.
GPT-OSS-20B has a smaller MoE structure, with 24 layers, 32 experts, and 4
active experts per token~\citep{openai2025gptoss}, so the same intervention has
less expert-mode redundancy to exploit before it removes functional diversity
(Figure~\ref{fig:gpt_oss_downstream}).

\begin{figure}[ht]
  \centering
  \includegraphics[width=0.6\linewidth]{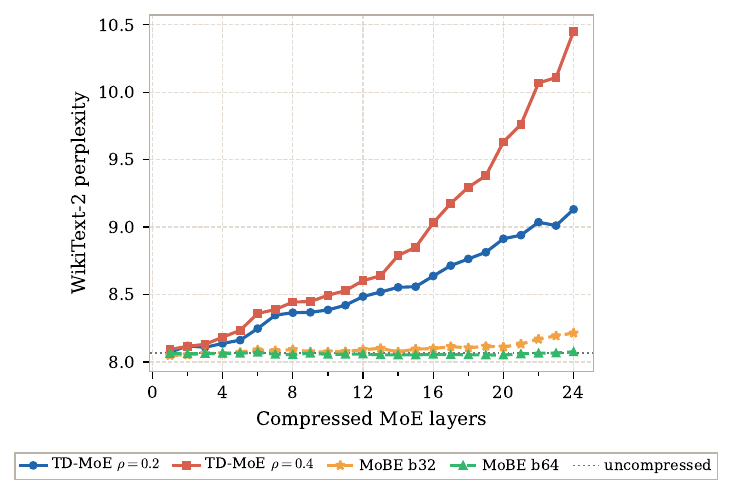}
  \caption{%
    WikiText-2 perplexity as the number of TD-MoE-compressed MoE layers increases on
    Qwen3-30B-A3B, for $\rho \in \{0.2, 0.4\}$. C4 perplexity is shown
    in the main text (Figure~\ref{fig:moe_main}).
  }
  \label{fig:qwen3_c4_ppl}
  \vspace{-8mm}
\end{figure}

\begin{figure*}[t]
  \centering
  \includegraphics[width=\linewidth]{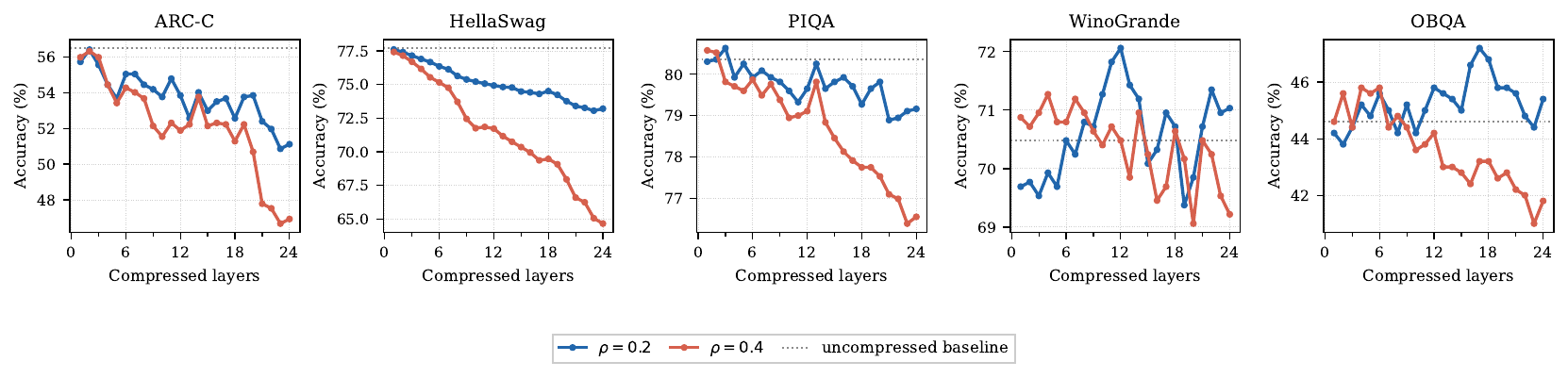}
  \caption{%
    Downstream task accuracy (\%) as the number of TD-MoE-compressed MoE
    layers increases on Qwen3-30B-A3B.
  }
  \label{fig:td_moe_downstream}
\end{figure*}

\begin{figure*}[t]
  \centering
  \includegraphics[width=\linewidth]{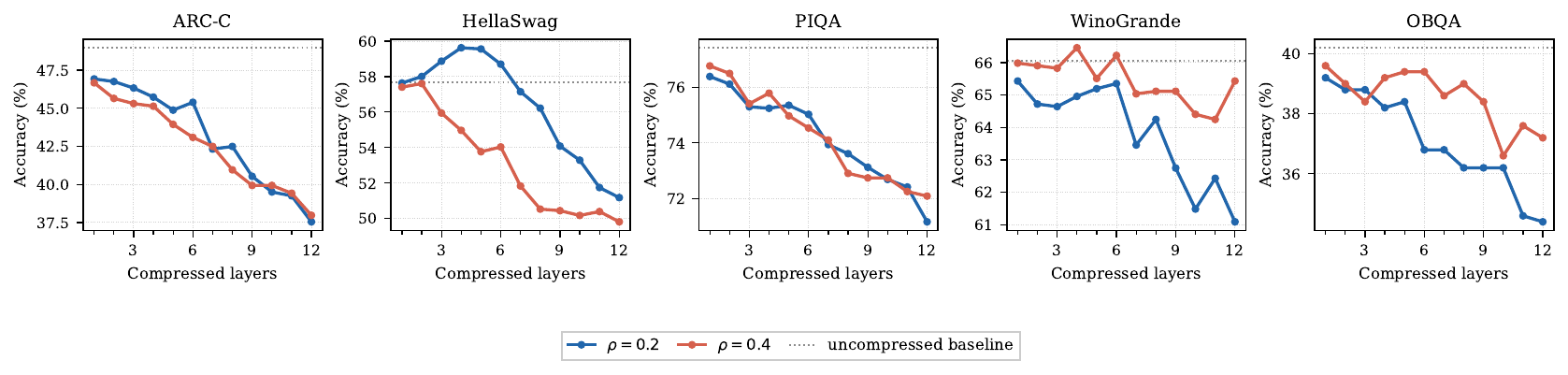}
  \vspace{-4mm}
  \caption{%
    Downstream task accuracy (\%) as the number of TD-MoE-compressed MoE
    layers increases on GPT-OSS-20B for $\rho \in \{0.2, 0.4\}$.
  }
  \label{fig:gpt_oss_downstream}
\end{figure*}

\begin{figure*}[h!]
  \centering
  \vspace{-4mm}\includegraphics[width=\linewidth]{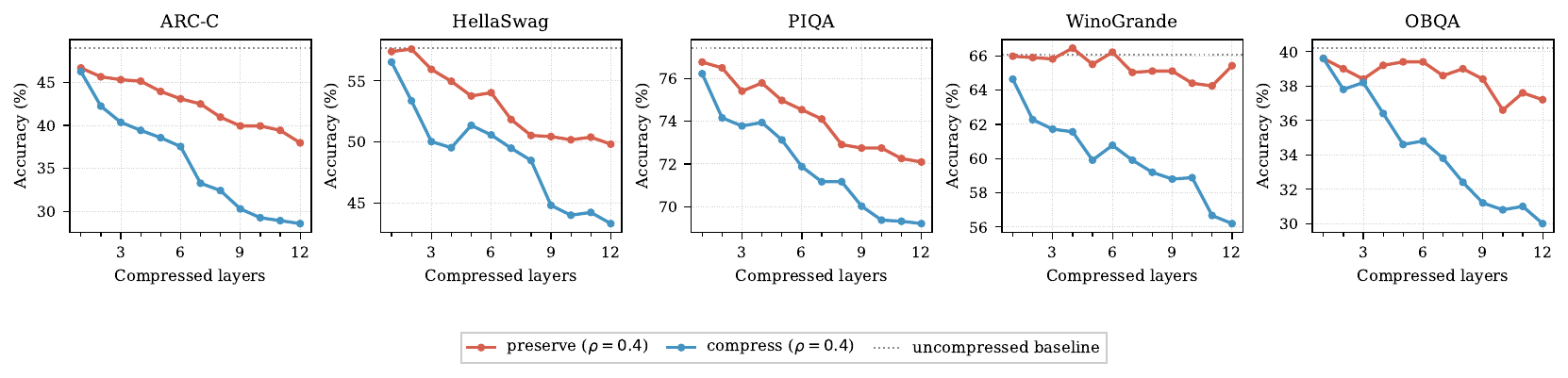}
  \caption{%
    Downstream accuracy (\%) versus number of TD-MoE-compressed layers on
    GPT-OSS-20B at $\rho{=}0.4$.
    \textsc{preserve} keeps the expert Tucker rank equal to $K$;
    \textsc{compress} additionally reduces the expert dimension.
  }
  \label{fig:gpt_oss_preserve_vs_compress}
  \vspace{-4mm}
\end{figure*}

\paragraph{Perplexity on GPT-OSS-20B.}
Unlike Qwen3-30B-A3B, GPT-OSS-20B reports anomalously high perplexity on raw
text: the uncompressed model scores $503.9$ on WikiText-2 and $269.9$ on C4.
This is a known artifact---GPT-OSS is built around the \emph{harmony} chat
format and is not intended for direct causal-language-model perplexity
evaluation on raw corpora, so its absolute perplexity is not a reliable quality signal. Consistent with
this, TD-MoE compression \emph{lowers} perplexity monotonically as more layers
are compressed (Figure~\ref{fig:gpt_oss_ppl}) rather than raising it, as the
intervention nudges the miscalibrated baseline toward more generic next-token
statistics. We therefore treat downstream accuracy
(Figure~\ref{fig:gpt_oss_downstream}) as the reliable signal for GPT-OSS-20B
and report perplexity only for completeness.



\subsection{Additional Experiments on GPT-OSS-20B}
\label{app:td_moe_gpt_oss}
\subsubsection{Expert-Mode Comparison}
We evaluate TD-MoE on OpenAI GPT-OSS-20B at per-layer compression
ratio $\rho{=}0.4$, varying the \emph{expert mode}: \textsc{preserve} fixes
the expert-dimension Tucker rank to $r_1{=}K$ (all experts are kept as
distinct latent directions), whereas \textsc{compress} also compresses the
expert dimension ($r_1 < K$). At the same target compression ratio, this
reallocates the saved expert-mode budget to the feature modes: in our
GPT-OSS-20B runs, \textsc{preserve} selects ranks $(32,1720,2664)$, while
\textsc{compress} selects $(20,2496,2880)$. Thus \textsc{compress} trades
expert-mode capacity for higher-rank input and output factors, allowing the
decomposition to exploit cross-expert redundancy at the cost of reduced expert
diversity.

Figure~\ref{fig:gpt_oss_preserve_vs_compress} reports the corresponding
downstream accuracy curves for \textsc{preserve} and \textsc{compress} as the
number of TD-MoE-compressed GPT-OSS-20B layers increases.

\subsubsection{Comparison Against MoBE}
\label{app:gpt_oss_mobe}

We extend the \texttt{MoBE} baseline introduced in Section~\ref{sec:td_moe_qwen} to GPT-OSS-20B. GPT-OSS-20B has only 32 experts per layer (against Qwen3-30B-A3B's 128), so MoBE's number of basis matrices
must satisfy $n_B < 32$ to yield any compression. We evaluate $n_B \in \{8, 16\}$, with the truncation $T$ fixed at the lowest per-expert dimension following MoBE's published convention.
The two settings are matched in bits-saved to TD-MoE $\rho{=}0.4$ and $\rho{=}0.2$ respectively, while all other protocol details (progressive middle-out block
schedule, evaluation datasets, activation-geometry probe) follow Appendix~\ref{app:td_moe}.

\begin{figure}[!b]
  \centering
  \vspace{-8mm}\includegraphics[width=0.8\linewidth]{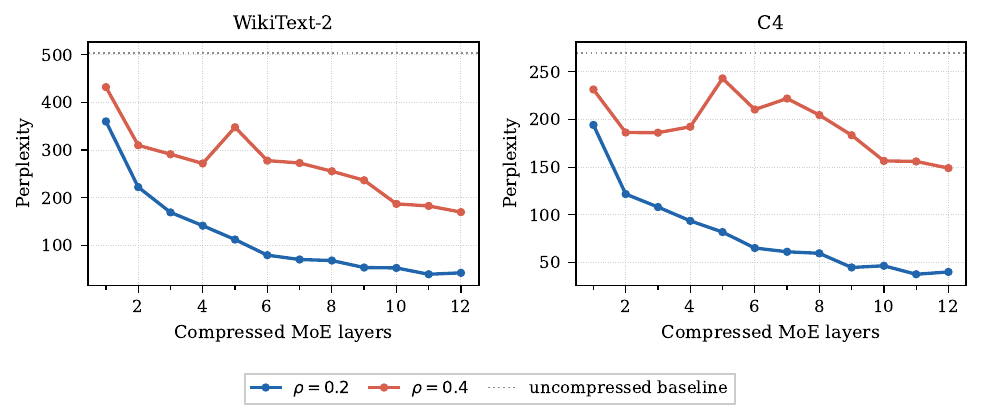}
  \caption{%
    WikiText-2 (left) and C4 (right) perplexity as the number of
    TD-MoE-compressed MoE layers increases on GPT-OSS-20B, for
    $\rho \in \{0.2, 0.4\}$. The uncompressed baseline (dotted) is anomalously
    high because GPT-OSS targets the harmony chat format rather than raw-text
    language modeling; perplexity is therefore not a reliable quality
    metric for this model.
  }
  \label{fig:gpt_oss_ppl}
  \vspace{-4mm}
\end{figure}

Figure~\ref{fig:gpt_oss_mobe_ppl} reports C4 perplexity as more MoE layers are compressed. As in the TD-MoE-only experiments (Figure~\ref{fig:gpt_oss_ppl}), absolute perplexity values on GPT-OSS-20B are not
interpretable in isolation because the uncompressed baseline is miscalibrated for raw-text causal-language-model evaluation. The relative ordering is nonetheless informative: MoBE perturbs the model less than
TD-MoE at matched bits-saved. We rely on the activation-geometry diagnostic in Figure~\ref{fig:gpt_oss_mobe_geometry} and downstream accuracy (Figure~\ref{fig:gpt_oss_downstream}) as the reliable
quality signals.

Figure~\ref{fig:gpt_oss_mobe_geometry} compares the methods at the residual-stream level. Each marker is one block-schedule setting; marker shape encodes the decomposition family, marker area is the fraction
of bits saved relative to the dense model, and colour encodes C4 perplexity.

\section{Computational Costs}

All model experiments were run on NVIDIA H100 GPUs. The GPT-J~6B and LLaMA~2~7B tensor-decomposition experiments described in Appendix~\ref{app:gptj_llama_tensor_protocol}, including the progressive
middle-layer compression grid, perplexity evaluation, activation-geometry diagnostics, and the lightweight LoRA repair runs, required approximately $100$ H100 GPU-hours in total. The Qwen3-30B-A3B and GPT-OSS-20B tensor-decomposition experiments described in Appendix~\ref{app:td_moe}, including the progressive
middle-layer compression grid, perplexity evaluation, activation-geometry diagnostics, and the MoBE comparison, required approximately $60$ H100 GPU-hours in total. We report GPU-hours as wall-clock runtime multiplied by the number of GPUs used.

All models (GPT-J 6B, LLaMA 2 7B, Qwen3-30B-A3B, GPT-OSS-20B) and benchmarks (WikiText-2, C4, ARC-Challenge, HellaSwag, OpenBookQA, PIQA, WinoGrande) are used strictly for research evaluation of compression methods, consistent with their publicly stated intended research use. We do not redistribute model weights and do not derive products outside research contexts

\begin{figure}[htbp]
\centering
\begin{minipage}[t]{0.46\linewidth}
    \centering
    \includegraphics[width=\linewidth]{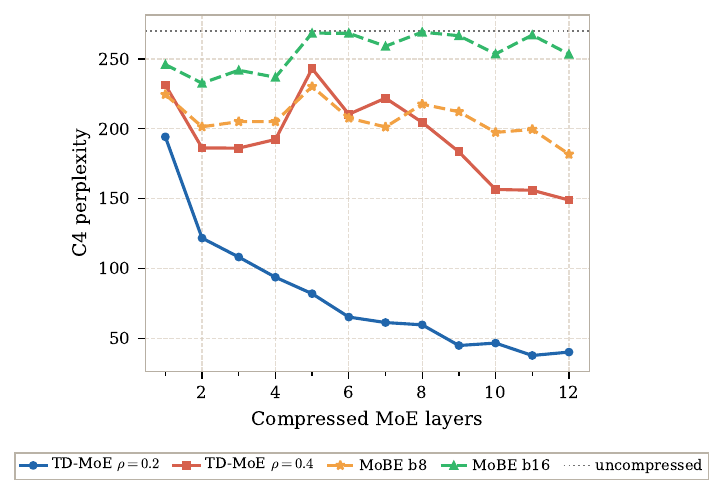}
    \captionof{figure}{%
        C4 perplexity as the number of compressed MoE layers increases on
        GPT-OSS-20B, for TD-MoE $\rho \in \{0.2, 0.4\}$ and MoBE
        $n_B \in \{8, 16\}$. The uncompressed C4 baseline (dotted) is
        anomalously high because GPT-OSS targets the harmony chat format
        rather than raw-text language modelling
        (cf.\ Figure~\ref{fig:gpt_oss_ppl}); absolute perplexity is not a
        reliable quality metric for this model, but the relative
        ordering---MoBE perturbing the model less than TD-MoE at matched
        bits-saved---is consistent with the activation-geometry diagnostic in
        Figure~\ref{fig:gpt_oss_mobe_geometry}.%
    }
    \label{fig:gpt_oss_mobe_ppl}
\end{minipage}\hfill%
\begin{minipage}[t]{0.52\linewidth}
    \centering
    \includegraphics[width=\linewidth]{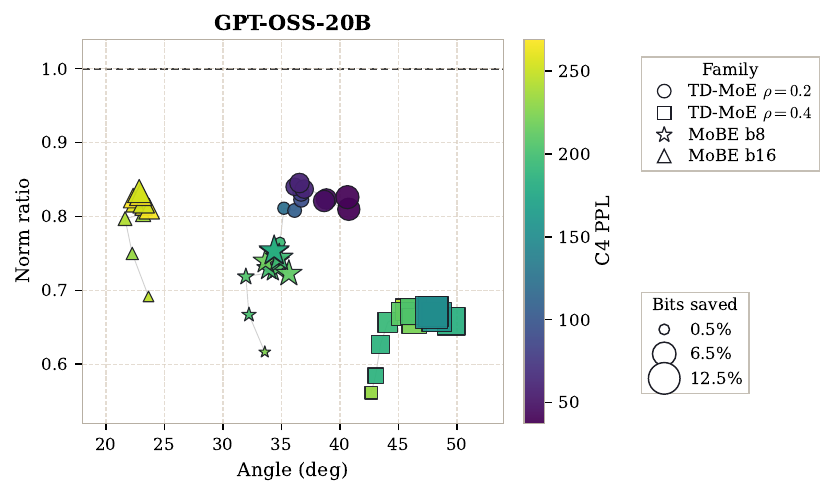}
    \captionof{figure}{%
        Residual-stream activation geometry of TD-MoE versus MoBE on
        GPT-OSS-20B. Each marker is one block-schedule setting; marker shape
        indicates the decomposition family, marker area is the fraction of
        bits saved relative to the dense model, and colour encodes C4
        perplexity. The dashed line marks the no-perturbation reference
        (norm ratio $=1$). At matched bits-saved, MoBE produces smaller
        MoE-output angles and norm ratios closer to one than TD-MoE,
        mirroring the Qwen3-30B-A3B finding (Figure~\ref{fig:moe_main}).%
    }
    \label{fig:gpt_oss_mobe_geometry}
\end{minipage}
\end{figure}



\end{appendixpart}

@inproceedings{unreasonable_ineffectiveness_2024,
  title={The unreasonable ineffectiveness of the deeper layers},
  author={Gromov, Andrey and Tirumala, Kushal and Shapourian, Hassan and Glorioso, Paolo and Roberts, Dan},
  year={2024},
  booktitle={The Thirteenth International Conference on Learning Representations}
}

@article{LLM_pruner_2023,
  title={Llm-pruner: On the structural pruning of large language models},
  author={Ma, Xinyin and Fang, Gongfan and Wang, Xinchao},
  journal={Advances in neural information processing systems},
  volume={36},
  pages={21702--21720},
  year={2023}
}

@article{OBC_2022,
  title={Optimal brain compression: A framework for accurate post-training quantization and pruning},
  author={Frantar, Elias and Alistarh, Dan},
  journal={Advances in Neural Information Processing Systems},
  volume={35},
  pages={4475--4488},
  year={2022}
}

@article{parkina2026coala,
  title={Coala: Numerically stable and efficient framework for context-aware low-rank approximation},
  author={Parkina, Uliana and Rakhuba, Maxim},
  journal={Advances in Neural Information Processing Systems},
  volume={38},
  pages={71014--71041},
  year={2026}
}

@article{GPTQ_2022,
  title={Gptq: Accurate post-training quantization for generative pre-trained transformers},
  author={Frantar, Elias and Ashkboos, Saleh and Hoefler, Torsten and Alistarh, Dan},
  journal={arXiv preprint arXiv:2210.17323},
  year={2022}
}

@article{AWQ_2024,
  title={Awq: Activation-aware weight quantization for on-device llm compression and acceleration},
  author={Lin, Ji and Tang, Jiaming and Tang, Haotian and Yang, Shang and Chen, Wei-Ming and Wang, Wei-Chen and Xiao, Guangxuan and Dang, Xingyu and Gan, Chuang and Han, Song},
  journal={Proceedings of machine learning and systems},
  volume={6},
  pages={87--100},
  year={2024}
}

@article{AQLM_2024,
  title={Extreme compression of large language models via additive quantization},
  author={Egiazarian, Vage and Panferov, Andrei and Kuznedelev, Denis and Frantar, Elias and Babenko, Artem and Alistarh, Dan},
  journal={arXiv preprint arXiv:2401.06118},
  year={2024}
}

@article{QUIP_2023,
  title={Quip: 2-bit quantization of large language models with guarantees},
  author={Chee, Jerry and Cai, Yaohui and Kuleshov, Volodymyr and De Sa, Christopher M},
  journal={Advances in neural information processing systems},
  volume={36},
  pages={4396--4429},
  year={2023}
}

@article{HintonDistil_2015,
  title={Distilling the knowledge in a neural network},
  author={Hinton, Geoffrey and Vinyals, Oriol and Dean, Jeff},
  journal={arXiv preprint arXiv:1503.02531},
  year={2015}
}

@inproceedings{GKD_2023,
  title={Gkd: A general knowledge distillation framework for large-scale pre-trained language model},
  author={Tan, Shicheng and Tam, Weng Lam and Wang, Yuanchun and Gong, Wenwen and Zhao, Shu and Zhang, Peng and Tang, Jie},
  booktitle={Proceedings of the 61st Annual Meeting of the Association for Computational Linguistics (Volume 5: Industry Track)},
  pages={134--148},
  year={2023}
}

@inproceedings{LASER_2024,
  title={The truth is in there: Improving reasoning in language models with layer-selective rank reduction},
  author={Sharma, Pratyusha and Ash, Jordan and Misra, Dipendra Kumar},
  booktitle={International Conference on Learning Representations},
  volume={2024},
  pages={17632--17651},
  year={2024}
}

@article{Attention_Is_All_You_Need_2017,
  title={Attention is all you need},
  author={Vaswani, Ashish and Shazeer, Noam and Parmar, Niki and Uszkoreit, Jakob and Jones, Llion and Gomez, Aidan N and Kaiser, {\L}ukasz and Polosukhin, Illia},
  journal={Advances in neural information processing systems},
  volume={30},
  year={2017}
}

@article{Eckart_Young_1936,
  title={The approximation of one matrix by another of lower rank},
  author={Eckart, Carl and Young, Gale},
  journal={Psychometrika},
  volume={1},
  number={3},
  pages={211--218},
  year={1936},
  publisher={Springer-Verlag}
}

@article{Eckart_Young_Mirsky_1960,
  title={Symmetric gauge functions and unitarily invariant norms},
  author={Mirsky, Leon},
  journal={The quarterly journal of mathematics},
  volume={11},
  number={1},
  pages={50--59},
  year={1960},
  publisher={Oxford University Press}
}

@article{MoE_2022,
  title={Switch transformers: Scaling to trillion parameter models with simple and efficient sparsity},
  author={Fedus, William and Zoph, Barret and Shazeer, Noam},
  journal={Journal of Machine Learning Research},
  volume={23},
  number={120},
  pages={1--39},
  year={2022}
}

@misc{li2026lestd,
  title={{LeSTD}: Learning Sparse {Tucker} Decomposition for Efficient Large Language Models},
  author={Yi Li and Zhichun Guo and Miao Yin, Bingzhe Li},
  howpublished={arXiv preprint arXiv:2601.01123},
  year={2026}
}

@inproceedings{FFNs_Are_Memory_2021,
  title={Transformer feed-forward layers are key-value memories},
  author={Geva, Mor and Schuster, Roei and Berant, Jonathan and Levy, Omer},
  booktitle={Proceedings of the 2021 Conference on Empirical Methods in Natural Language Processing},
  pages={5484--5495},
  year={2021}
}

@article{SinksAndSpikes_2026,
  title={The spike, the sparse and the sink: Anatomy of massive activations and attention sinks},
  author={Sun, Shangwen and Canziani, Alfredo and LeCun, Yann and Zhu, Jiachen},
  journal={arXiv preprint arXiv:2603.05498},
  year={2026}
}

@article{touvron2023llama,
  title={Llama 2: Open foundation and fine-tuned chat models},
  author={Touvron, Hugo and Martin, Louis and Stone, Pierre and Albert, Benjamin and Almahairi, Amjad and Babaei, Yasmine and Bashlykov, Nikolay and Batra, Soumya and Bhargava, Prajwal and Bhosale, Shruti and others},
  journal={arXiv preprint arXiv:2307.09288},
  year={2023}
}

@misc{gpt-j,
  author = {Wang, Ben and Komatsuzaki, Aran},
  title = {{GPT-J-6B: A 6 Billion Parameter Autoregressive Language Model}},
  howpublished = {\url{https://github.com/kingoflolz/mesh-transformer-jax}},
  year = 2021,
  month = May
}

@article{tucker1966some,
  title={Some Mathematical Notes on Three-Mode Factor Analysis},
  author={Tucker, Ledyard R.},
  journal={Psychometrika},
  volume={31},
  number={3},
  pages={279--311},
  year={1966}
}

@article{kolda2009tensor,
  title={{Tensor Decompositions and Applications}},
  author={Kolda, Tamara G. and Bader, Brett W.},
  journal={SIAM Review},
  volume={51},
  number={3},
  pages={455--500},
  year={2009}
}

@article{oseledets2011tensor,
  title={Tensor-Train Decomposition},
  author={Oseledets, Ivan V.},
  journal={SIAM Journal on Scientific Computing},
  volume={33},
  number={5},
  pages={2295--2317},
  year={2011}
}

@misc{luo2024trawl,
  title={{TRAWL}: Tensor Reduced and Approximated Weights for Large Language Models},
  author={Luo, Yiran and Patel, Het and Fu, Yu and Ahn, Dawon and Chen, Jia and Dong, Yue and Papalexakis, Evangelos E.},
  howpublished={arXiv preprint arXiv:2406.17261},
  year={2024}
}

@inproceedings{huang2025sola,
  title     = {{SoLA}: Leveraging Soft Activation Sparsity and Low-Rank
               Decomposition for Large Language Model Compression},
  author    = {Huang, Xinhao and Huang, You-Liang and Wen, Zeyi},
  booktitle = {Proceedings of the {AAAI} Conference on Artificial Intelligence},
  volume    = {39},
  number    = {16},
  pages     = {17494--17502},
  year      = {2025},
  doi       = {10.1609/aaai.v39i16.33923}
}

@inproceedings{tian2025flatllm,
  title     = {{FLAT-LLM}: Fine-Grained Low-Rank Activation Space
               Transformation for Large Language Model Compression},
  author    = {Tian, Jiayi and others},
  booktitle = {Findings of the Association for Computational Linguistics:
               {EACL} 2026},
  year      = {2026},
  eprint    = {2505.23966},
  archivePrefix = {arXiv}
}

@InProceedings{tensorllm2025,
  title={TensorLLM: Tensorising Multi-Head Attention for Enhanced Reasoning and Compression in LLMs},
  author={Gu, Yuxuan and Zhou, Wuyang and Iacovides, Giorgos and Mandic, Danilo},
  booktitle={International Joint Conference on Neural Networks (IJCNN)},
  year={2025},
  url={https://arxiv.org/abs/2501.15674},
  eprint={2501.15674}
}

@InProceedings{tdmoe2026,
  title={TD-MoE: Tensor Decomposition for MoE Models},
  author={Xu, Yuebin and Wang, Yanhong and Peng, Xuemei and Zang, Hui and Chen, Minghao and Xia, Pengfei and Wen, Zeyi},
  booktitle={International Conference on Learning Representations (ICLR)},
  year={2026},
  url={https://openreview.net/forum?id=D9cnZNZfxX},
  note={ICLR 2026}
}

@misc{anjum2024ttlora,
  title={Tensor Train Low-rank Approximation ({TT-LoRA}): Democratizing {AI} with Accelerated {LLMs}},
  author={Anjum, Afia and Eren, Maksim E. and Boureima, Ismael and Alexandrov, Boian and Bhattarai, Manish},
  howpublished={arXiv preprint arXiv:2408.01008},
  year={2024},
  url={https://arxiv.org/abs/2408.01008}
}

@misc{yang2024adazeta,
  title={{AdaZeta}: Adaptive Zeroth-Order Tensor-Train Adaption for Memory-Efficient Large Language Models Fine-Tuning},
  author={Yang, Yifan and Zhen, Kai and Banijamal, Ershad and Mouchtaris, Athanasios and Zhang, Zheng},
  howpublished={arXiv preprint arXiv:2406.18060},
  year={2024},
  url={https://arxiv.org/abs/2406.18060},
  note={Accepted to EMNLP 2024}
}

@misc{yang2024loretta,
  title={{LoRETTA}: Low-Rank Economic Tensor-Train Adaptation for Ultra-Low-Parameter Fine-Tuning of Large Language Models},
  author={Yang, Yifan and Zhou, Jiajun and Wong, Ngai and Zhang, Zheng},
  howpublished={arXiv preprint arXiv:2402.11417},
  year={2024},
  url={https://arxiv.org/abs/2402.11417}
}

@article{sun2024massive,
  title={Massive Activations in Large Language Models}, 
  author={Sun, Mingjie and Chen, Xinlei and Kolter, J. Zico and Liu, Zhuang},
  year={2024},
  journal={arXiv preprint arXiv:2402.17762}
}

@InProceedings{svdllm2025,
  title={SVD-LLM: Truncation-aware Singular Value Decomposition for Large Language Model Compression},
  author={Wang, Xinyi and Yuan, Zhihang and Wang, Yuang and Yuan, Qiang and Sun, Guangyu and Zhou, Weiyang},
  booktitle={International Conference on Learning Representations (ICLR)},
  year={2025},
  url={https://arxiv.org/abs/2403.07378},
  eprint={2403.07378}
}

@article{wang2025dobisvd,
  title={Dobi-SVD: Differentiable SVD for LLM Compression and Some New Perspectives},
  author={Wang, Qinsi and Ke, Jinghan and Tomizuka, Masayoshi and Chen, Yiran and Keutzer, Kurt and Xu, Chenfeng},
  journal={arXiv preprint arXiv:2502.02723},
  year={2025},
  url={https://arxiv.org/abs/2502.02723},
  eprint={2502.02723}
}

@article{hasslefree2025,
  title={HASSLE-free: A Unified Framework for Sparse plus Low-Rank Matrix Decomposition for LLMs},
  author={Makni, Mehdi and Behdin, Kayhan and Xu, Zheng and Ponomareva, Natalia and Mazumder, Rahul},
  journal={arXiv preprint arXiv:2502.00899},
  year={2025},
  url={https://arxiv.org/abs/2502.00899},
  eprint={2502.00899}
}

@article{slicegpt2024,
  title={SliceGPT: Compress Large Language Models by Deleting Rows and Columns},
  author={Ashkboos, Saleh and Croci, Maximilian L and do Nascimento, Marcelo Gennari and Hoefler, Torsten and Hensman, James},
  journal={arXiv preprint},
  year={2024},
  url={https://arxiv.org/abs/2401.15024},
  eprint={2401.15024}
}

@misc{openai2025gptoss,
  title={Introducing {gpt-oss}},
  author={{OpenAI}},
  year={2025},
  howpublished={\url{https://openai.com/index/introducing-gpt-oss/}},
  note={Accessed: 2026-05-08}
}

@article{su2025superexperts,
  title={Unveiling Super Experts in {Mixture-of-Experts} Large Language Models},
  author={Su, Zunhai and Li, Qingyuan and Zhang, Hao and Qian, YuLei and Xie, Yuchen and Yuan, Kehong},
  journal={arXiv preprint arXiv:2507.23279},
  year={2025},
  url={https://arxiv.org/abs/2507.23279},
  eprint={2507.23279}
}

@article{qwen3technicalreport,
  title={Qwen3 technical report},
  author={Yang, An and Li, Anfeng and Yang, Baosong and Zhang, Beichen and Hui, Binyuan and Zheng, Bo and Yu, Bowen and Gao, Chang and Huang, Chengen and Lv, Chenxu and others},
  journal={arXiv preprint arXiv:2505.09388},
  year={2025}
}

@article{superweight,
  title={The super weight in large language models},
  author={Yu, Mengxia and Wang, De and Shan, Qi and Reed, Colorado J and Wan, Alvin},
  journal={arXiv preprint arXiv:2411.07191},
  year={2024}
}

@article{chen2025mobe,
  title={MoBE: Mixture-of-Basis-Experts for Compressing MoE-based LLMs},
  author={Chen, Xiaodong and Ha, Mingming and Lan, Zhenzhong and Zhang, Jing and Li, Jianguo},
  journal={arXiv preprint arXiv:2508.05257},
  year={2025}
}

@article{hosvdopt,
  title={A multilinear singular value decomposition},
  author={De Lathauwer, Lieven and De Moor, Bart and Vandewalle, Joos},
  journal={SIAM journal on Matrix Analysis and Applications},
  volume={21},
  number={4},
  pages={1253--1278},
  year={2000},
  publisher={SIAM}
}

@article{martin2021implicit,
  title={Implicit self-regularization in deep neural networks: Evidence from random matrix theory and implications for learning},
  author={Martin, Charles H and Mahoney, Michael W},
  journal={Journal of Machine Learning Research},
  volume={22},
  number={165},
  pages={1--73},
  year={2021}
}

@article{friedland2018nuclear,
  title={Nuclear norm of higher-order tensors},
  author={Friedland, Shmuel and Lim, Lek-Heng},
  journal={Mathematics of Computation},
  volume={87},
  number={311},
  pages={1255--1281},
  year={2018}
}

@article{tyrtyshnikov2003tensor,
  title={Tensor approximations of matrices generated by asymptotically smooth functions},
  author={Tyrtyshnikov, Eugene Evgen'evich},
  journal={Sbornik: Mathematics},
  volume={194},
  number={6},
  pages={941--954},
  year={2003}
}

@article{hackbusch2006low,
  title={Low-rank Kronecker-product approximation to multi-dimensional nonlocal operators. Part II. HKT representation of certain operators},
  author={Hackbusch, Wolfgang and Khoromskij, Boris N},
  journal={Computing},
  volume={76},
  number={3},
  pages={203--225},
  year={2006},
  publisher={Springer}
}

@incollection{gholami2022survey,
  title={A survey of quantization methods for efficient neural network inference},
  author={Gholami, Amir and Kim, Sehoon and Dong, Zhen and Yao, Zhewei and Mahoney, Michael W and Keutzer, Kurt},
  booktitle={Low-power computer vision},
  pages={291--326},
  year={2022},
  publisher={Chapman and Hall/CRC}
}

@article{harp,
      title={HARP: Hadamard-Preconditioned Adaptive Rotation Processor for Extreme LLM Quantization}, 
      author={Artur Zagitov and Gleb Molodtsov and Aleksandr Beznosikov},
      year={2026},
      eprint={2605.29843},
      archivePrefix={arXiv},
      primaryClass={cs.LG},
      url={https://arxiv.org/abs/2605.29843}, 
}

@article{switch,
  title={Switch transformers: Scaling to trillion parameter models with simple and efficient sparsity},
  author={Fedus, William and Zoph, Barret and Shazeer, Noam},
  journal={Journal of Machine Learning Research},
  volume={23},
  number={120},
  pages={1--39},
  year={2022}
}

@article{moge,
  title={Pangu pro moe: Mixture of grouped experts for efficient sparsity},
  author={Tang, Yehui and Li, Xiaosong and Liu, Fangcheng and Guo, Wei and Zhou, Hang and Wang, Yaoyuan and Han, Kai and Yu, Xianzhi and Li, Jinpeng and Zang, Hui and others},
  journal={arXiv preprint arXiv:2505.21411},
  year={2025}
}

@inproceedings{himoe,
  title={Hierarchical Mixture-of-Experts with Two-Stage Optimization},
  author={Molodtsov, Gleb and Miasnikov, Alexander and Beznosikov, Aleksandr},
  booktitle={ICML 2026 Workshop on Weight-Space Symmetries: from Foundations to Practical Applications},
  year={2026}
}
\end{document}